\documentclass[conference]{IEEEtran}
\IEEEoverridecommandlockouts
\usepackage{graphicx}
\usepackage{amsmath}
\usepackage{amssymb}
\usepackage{booktabs}
\usepackage{mathtools}
\usepackage{tabularx}
\usepackage{caption}
\usepackage{xspace}
\usepackage{subcaption} 

\usepackage[pagebackref,breaklinks,colorlinks]{hyperref}

\usepackage[capitalize]{cleveref}
\crefname{section}{Sec.}{Secs.}
\Crefname{section}{Section}{Sections}
\Crefname{table}{Table}{Tables}
\crefname{table}{Tab.}{Tabs.}

\def\BibTeX{{\rm B\kern-.05em{\sc i\kern-.025em b}\kern-.08em
    T\kern-.1667em\lower.7ex\hbox{E}\kern-.125emX}}
\begin{document}

\title{Visual Concept-driven Image Generation with Text-to-Image Diffusion Model\\
%{\footnotesize \textsuperscript{*}Note: Sub-titles should not be used}
%\thanks{Identify applicable funding agency here. If none, delete this.}
}

\author{Tanzila Rahman$^{1,3}$ \qquad  Shweta Mahajan$^{1,3}$ \qquad Hsin-Ying Lee$^{2}$ \qquad Jian Ren$^{2}$ \\ 
\qquad Sergey Tulyakov$^{2}$  \qquad Leonid Sigal$^{1,3,4}$\\
\textit{$^1$University of British Columbia} \qquad \textit{$^2$Snap Inc.} \\
\textit{$^3$Vector Institute for AI} \qquad
\textit{$^4$Canada CIFAR AI Chair} \\
%\texttt{trahman8@cs.ubc.ca, my.yang@mail.utoronto.ca,  lsigal@cs.ubc.ca}
}

%\maketitle

%%%%%%%%% ABSTRACT
\makeatletter
\DeclareRobustCommand\onedot{\futurelet\@let@token\@onedot}
\def\@onedot{\ifx\@let@token.\else.\null\fi\xspace}

\def\eg{\emph{e.g}\onedot} \def\Eg{\emph{E.g}\onedot}
\def\ie{\emph{i.e}\onedot} \def\Ie{\emph{I.e}\onedot}
\def\cf{\emph{cf}\onedot} \def\Cf{\emph{C.f}\onedot}
\def\etc{\emph{etc}\onedot} \def\vs{\emph{vs}\onedot}
\def\wrt{w.r.t\onedot} \def\dof{d.o.f\onedot}
\def\etal{\emph{et al}\onedot}
\makeatother

\twocolumn[{
\renewcommand\twocolumn[1][]{#1}
\vspace*{-10mm}
\maketitle

\begin{center}
    \centering
    % \fbox{\rule{0pt}{2in} \rule{0.9\linewidth}{0pt}}
    \includegraphics[width=\textwidth]{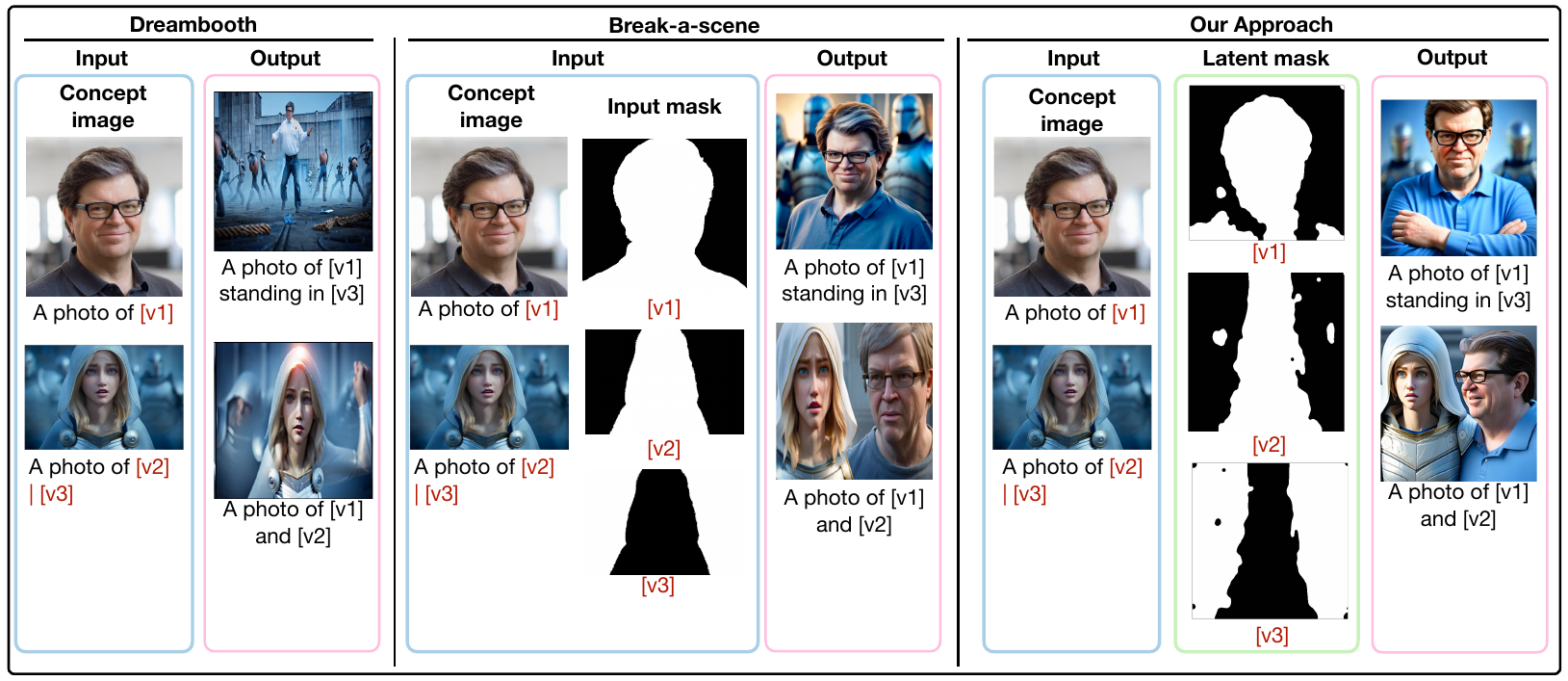}
    % \vspace*{5mm}
    %\vspace{-0.2in}
    \captionof{figure}{{\bf Concept-driven image generation:} Given images depicting multiple concepts (\emph{subjects and context/background}), 
the top output is the illustration of the male ({\small \tt [v1]]}) being generated in the context/background of the female ({\small \tt [v3]}) image by different methods.
The bottom output illustrates the two concepts together in a single image. 
Dreambooth \cite{ruiz2023dreambooth} (left) encodes \texttt{[v1], [v2], [v3]} from multiple input concept images. It fails to generate multi-concept interactions.
Break-a-scene \cite{avrahami2023break} (middle) disentangles \texttt{[v1], [v2], [v3]} from single image. This approach requires human-annotated masks.
Our approach (right) disentangles \texttt{[v1], [v2], [v3]} from single image.
The \emph{latent masks} are obtained from EM-like optimization. 
Using these optimized masks, our method can automatically produce images with those concepts in new contexts either by themselves or, jointly, interacting with one another.
    }
    \label{fig1:intro}
\end{center}
}]

\begin{abstract}
Text-to-image (TTI) diffusion models have demonstrated impressive results in generating high-resolution images of complex and imaginative scenes. Recent approaches have further extended these methods with personalization techniques that allow them to integrate user-illustrated concepts (e.g., the user him/herself) using a few sample image illustrations. However, the ability to generate images with multiple interacting concepts, such as human subjects, as well as concepts that may be entangled in one, or across multiple, image illustrations remains illusive. In this work, we propose a concept-driven TTI personalization framework that addresses these core challenges. We build on existing works that learn custom tokens for user-illustrated concepts, allowing those to interact with existing text tokens in the TTI model. However, importantly, to disentangle and better learn the concepts in question, we jointly learn (latent) segmentation masks that disentangle these concepts in user-provided image illustrations. We do so by introducing an Expectation Maximization (EM)-like optimization procedure where we alternate between learning the custom tokens and estimating (latent) masks encompassing corresponding concepts in user-supplied images. We obtain these masks based on cross-attention, from within the U-Net parameterized latent diffusion model and subsequent DenseCRF optimization. We illustrate that such joint alternating refinement leads to the learning of better tokens for concepts and, as a by-product, latent masks. We illustrate the benefits of the proposed approach qualitatively and quantitatively with several examples and use cases that can combine three or more entangled concepts. 
\end{abstract}


%%%%%%%%% BODY TEXT

%\vspace{-0.30in}
\section{Introduction}

\label{sec:intro}
Photorealistic image synthesis, particularly with text-to-image diffusion models ({\em e.g.}, Stable Diffusion \cite{rombach2021highresolution}), has provided a great boost to content creation.
Images of diverse and imaginative scenarios such as `` {\em A dog in space}'' can now be easily generated. %  with these models.
While such approaches can faithfully generate many instances of the generic subject, \eg, a {\em dog}, in a variety of scenarios, \eg, {\em space}, they are limited when the goal is to render a specific subject, 
\eg, my Beagle dog Bailey with unique fur markings, in a specific 
context or background, \eg, next to my Tudor-style house.
Further, the generative capabilities fall apart when modeling multi-object scenarios and interactions, \eg, me playing with Bailey next to my house. 
Since these cases are frequent and typical in visual content creation, 
such as stories or videos, it is useful to consider how to make existing 
text-to-image diffusion models more controllable and capable of 
generating multiple interacting user-specified concepts.

To introduce additional controllability, for example, concerning subject appearance or style and their occurrence in different contexts, many recent approaches \cite{gal2022image, han2023highly, yang2023controllable,ruiz2023dreambooth} have focused on controlling generation with sample images depicting concepts of interest. 
% encoding subjects with specific styles and subsequently generating the subjects in different contexts. 
For example, to render a specific subject, these approaches present multiple (usually relatively clean) images of the target subject to extract the single subject-specific token across these images, \eg, using inversion techniques, and then introduce this special token in prompts in conjunction with the regular text to generate desired content. 
However, while one can generate single concept/subject scenarios in this way, composing \emph{multiple} stylistically diverse concepts remains a challenge.

Consider images in \cref{fig1:intro}; approaches discussed above are capable of extracting the specific subject (from the left concept images) and ``pasting'' them in different contexts. 
However, they have shown to be significantly limited \cite{avrahami2023break} when
asked to generate images where, for example, two characters in question interact or when multiple concepts come from the same image (\eg, the top-left image can be the source of both the {\em subject} and the {\em background}). 
Learning of tokens and subsequent generation, in this scenario, requires explicit \emph{disentanglement} of the characters and the background. 
Further, stylistic variations (top -- game rendering; bottom -- real photograph) present added challenges. 
% , especially if multiple concepts are illustrated in the same image.

To address these limitations, recent work \cite{avrahami2023break} proposes a two-stage training pipeline for extracting multiple concepts from a single image by leveraging a user-specified mask for each of the concepts during the optimization of tokens associated with these concepts.
This mechanism enables the decomposition of an image into constituent subjects and the background.
Analogous to prior methods that use multiple images to learn a single concept, this approach also learns how to encode features of each of the concepts into corresponding tokens.
A key feature of the pipeline lies in the fine-tuning of model weights while optimizing the prompts for different concepts simultaneously. 
This ensures that multiple concepts can be compositionally combined.
While this works well in practice, the process requires users to specify masks for concepts they desire, which requires a custom sketching interface and additional
user effort (as shown in \cref{fig1:intro},~middle). 
% \TODO[Does is work for out of domain]

In this work, we attempt to do away with the need to specify masks for multi-concept extraction and composition, while still leveraging (latent) masking to learn disentangled concepts that can then be composed using the text-to-image diffusion model~\cite{rombach2021highresolution}. 
Specifically, we utilize loose textual input prompts and the cross-attention maps for the target concept tokens to generate masks during prompt optimization. However, generating concept masks from cross-attention ({\em e.g.}, as is done in \cite{wu2023diffumask}) of pre-trained diffusion model tends to be noisy; often under- or over-segmenting the target concept(s). Therefore, we propose a joint EM-style optimization over the concept tokens and their (latent) masks, where given mask initializations (obtain in the style of \cite{wu2023diffumask} from cross-attention) we iterate between learning tokens for masked concepts and re-estimation of the mask itself. We illustrate that this leads to both learning of better concept tokens and, as a by-product, a better latent mask for the concepts in question. 

% In this work, we go a step further and consider the unique problem of replacing subjects in target images with new out-of-distribution subjects and even inserting the out-of-distribution subjects along with the in-domain subjects to generate novel images displaying rich interactions. 
% Supplementing each image with custom masks in this setting is not always feasible owing to the exclusivity of the concepts or designing these custom masks can be an expensive and time-consuming process.
% Therefore, instead of relying on the ground-truth masks for concept disentanglement, we utilize the input prompts and the cross-attention maps for the target concept tokens to generate masks during prompt optimization.

As is illustrated in \cref{fig1:intro} the proposed approach can automatically learn tokens corresponding to the two subjects ({\small \tt [v1], \tt [v2]})  and their backgrounds. Given these tokens, it can seamlessly integrate subject in {\small \tt [v3]} within the environment of {\small \tt [v2]} and can also generate plausible interactions between the two subjects. Note that it can do so despite stylistic differences between the concept sources. 
% 
% , our approach seamlessly integrates {\small \tt concept 2} 
% with the subject information of two unrelated subjects ({\small \tt concept 1} and {\small \tt concept 2}), our approach seamlessly integrates {\small \tt concept 2} in the environment of {\small \tt concept 1} (subject replacement) and can also introduce interactions between the two subjects in a specified context.
% Notably, contexts such as those corresponding to {\small \tt concept 1} can also be learned through our proposed optimization mechanism.
Our contributions are:
\begin{itemize}
    \item We address a unique and challenging problem of personalized multi-subject generation with complex interactions of (potentially unrelated) subjects. 

    \item We propose an Expectation Maximization (EM)-like optimization procedure to disentangle concepts from a single, or multiple, images by generating masks for specific target concepts. 
    Our approach performs token optimization while simultaneously optimizing corresponding (latent) masks such that the token embedding best represents the disentangled concept.

    \item We show through novel combinations of subjects and interactive environments that our approach works well in practice and can generate realistic scenarios with custom subjects interacting in complex scenes and environments.
\end{itemize}

%contribution:
%\begin{itemize}
 % \item Dreambooth could not interact multiple characters from different concepts. Break-a-scene overcomes that limitation but we need ground truth masks. Therefore, preparing the annotations is expensive and time consuming. 
%  \item In our method, we are not dependent on ground truth annotated masks, rather we generate masks conditioned on prompt from the attention map and optimize those masks over multiple time steps. In that case we are solving the problem in a weakly supervised way.
%  \item We are not introducing any additional training stage. However, our method could solve both mask generation and combining multiple concepts from multiple images simultaneously.
%\end{itemize}

%\vspace{0.10in}
\section{Related Work}
%\vspace{-0.35in}
\noindent
\paragraph{Image and text-to-image generation.}
Realistic and high-quality image synthesis is now possible with the outstanding advances in deep generative models.  
Generative Adversarial Networks (GANs)~\cite{gal2022image, goodfellow2014generative, li2019controllable,karras2020analyzing, karras2019style, brock2018large, abdal2019image2stylegan, abdal2020image2stylegan++, abdal2022clip2stylegan} facilitate the creation of high-fidelity images spanning different domains. 
In addition to GANs, Variational Autoencoders (VAEs)~\cite{kingma2019introduction, vahdat2020nvae} have presented a likelihood-based method for generating images. 
Important class of generative models include  autoregressive models ~\cite{van2016conditional, parmar2018image, esser2021taming,chen2020generative} and diffusion models~\cite{rombach2021highresolution, dhariwal2021diffusion, ho2022cascaded, ho2020denoising, nichol2021improved, chefer2023attend, saharia2022palette}. 
The former views image pixels as a sequence with pixel-by-pixel correlation, while the latter generates images by gradually removing noise. 
Methods such as DALL-E~\cite{ramesh2021zero} employed an autoregressive transformer~\cite{vaswani2017attention} trained on both text and image tokens, showcasing remarkable zero-shot capabilities. 

Diffusion-based models \cite{ho2020denoising} with their spectacular realistic synthesis capabilities have risen as the leading choice for generating images from text descriptions~\cite{gu2022vector, ramesh2022hierarchical, kim2022diffusionclip, kim2021diffusionclip, xie2023boxdiff, he2023data}.
Some notable works on conditional diffusion models include GLIDE~\cite{nichol2021glide} which applies classifier and classifier-free guidance strategies for high-fidelity image generation.
Methods such as LAFITE~\cite{zhou2111lafite} used a pre-trained CLIP~\cite{radford2021learning} model to create a unified space for both text and images, enabling the training of text-to-image models without relying on specific textual datasets.
While these approaches were applied directly in the high-dimensional space of images, recent work, Latent Diffusion Model (LDM) \cite{rombach2021highresolution} for text-to-image generation apply diffusion to a low-dimensional visual representation and have, thus,  reduced the computational overhead of the diffusion models applied in the image space.
In this work, we build on latent diffusion models.

%It drew inspiration from models generating high-quality images without specific conditions, employing guided inference techniques, including both classifier and non-classifier methods, to create high-fidelity images. LAFITE~\cite{zhou2111lafite} used a pre-trained CLIP~\cite{radford2021learning} model to create a unified space for both text and images, enabling the training of text-to-image models without relying on specific textual datasets. In conjunction with these advancements, there has been also a rise in text-powered image editing, allowing for both overall and precise modifications to images~\cite{brooks2023instructpix2pix, crowson2022vqgan, meng2021sdedit, patashnik2021styleclip, avrahami2023blended, bar2022text2live}. 

\noindent
%\vspace{-0.25in}
\paragraph{Personalizing image generation.} Personalization involves identifying a concept provided by the user that is not commonly found in the training data for discriminative~\cite{cohen2022my} or generative~\cite{nitzan2022mystyle} purposes.
Current approaches for customized text-to-image generation with diffusion models have focussed on two types of strategies.
In the first, methods \cite{ruiz2023dreambooth, kumari2023multi, han2023svdiff, fei2023gradient, chen2023disenbooth, ma2023subject} adopt test-time fine-tuning where model weights are updated using a collection of images representative of the target concept/subject for personalization.
%At present, there exist two primary frameworks for customized text-to-image generation, distinguished by their approach to test-time fine-tuning. 
%\leon{[I am failing to see what the two classes are and how they are distinguished :/ Is that that some models optimize the model itself and others fix model and optimize tokens?]}
%Within the category of test-time fine-tuning strategies, some methods involve utilizing multiple personalized images centered around a particular subject . 
%These approaches directly adjust the model using these subject-specific images for fine-tuning  
%\leon{[Past this point, things are good]}
The second strategy is to focus solely on refining the token embedding representing the subject while keeping the model weights fixed to enhance the model's comprehension of visual concepts~\cite{gal2022image, han2023highly, yang2023controllable}.
In this direction, DreamBooth~\cite{ruiz2023dreambooth} takes a holistic approach by fine-tuning the entirety of the UNet network. 
In contrast, Custom Diffusion~\cite{kumari2023multi} concentrates its fine-tuning efforts exclusively on the K and V layers within the UNet network's cross-attention mechanism. This fine-tuning process is then optimized using LoRA~\cite{hu2021lora}. 
%Notably, Custom Diffusion introduces a groundbreaking technique for personalized generation across various subjects. 
Conversely, SVDiff~\cite{han2023svdiff} pioneers the use of cut mix to construct training data and introduces regularization penalties to mitigate overlapping attention maps associated with multiple subjects during the training process. Cones~\cite{liu2023cones} introduces the notion of concept neurons and exclusively modifies these neurons related to a specific subject within the K and V layers of cross-attention. 
To generate multiple personalized subjects, Cones directly combines the concept neurons from several trained personalized models. 
In contrast, Mix-of-Show~\cite{gu2023mix} adopts a different approach by training individual LoRA models for each subject and subsequently merging their outputs through fusion techniques.
In this work, we explore the integration of subjects extracted from multiple images, examining their mutual influence to generate images driven by these subjects. We utilize cross-attention maps to disentangle acquired subjects by optimizing masks.

\noindent
%\vspace{-0.25in}
\paragraph{Inversion in generative models.} Inversion tasks in generative models involve identifying a latent code, typically within the latent space of a generative model, that can faithfully recreate a specific image~\cite{zhu2016generative, xia2022gan}. The process of inversion is achieved either through optimization-based methods that directly refine a latent vector ~\cite{abdal2019image2stylegan, zhu2020improved, gu2020image} or by employing an encoder to discern and generate the latent representation corresponding to a given image~\cite{richardson2021encoding, zhu2020domain, tov2021designing}. In this work, we adhere to the optimization method, considering its superior adaptability to novel or unfamiliar concepts. The work incorporates a dual-phase strategy. Initially, we optimize solely the textual embeddings related to the target concepts using masks generated by cross-attention maps. Subsequently, we undertake joint training, refining both the embeddings through mask optimization and the model weights simultaneously.

\section{Methodology}

\begin{figure*}[t!]
  \centering
  \includegraphics[scale=0.55]{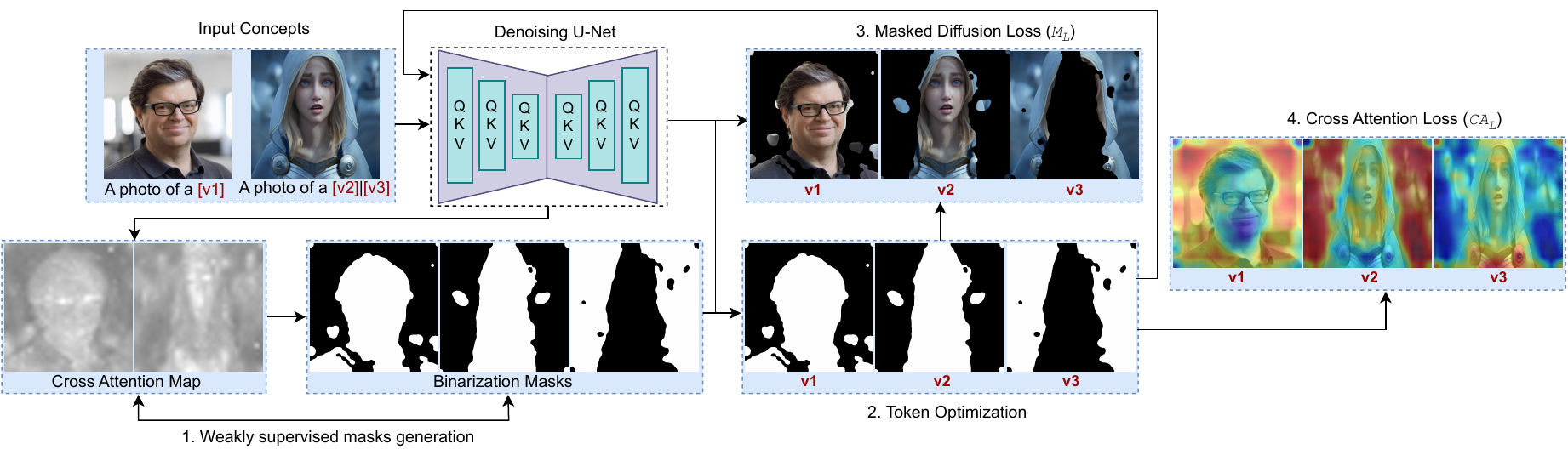}
%\vspace{-0.1in}
\caption{
{\bf Overview of our EM-style Optimization Framework.} 
Unique tokens are first assigned to the concepts. In this example, three tokens are assigned: {\tt [v1]} to the male subject in the first image, {\tt [v2]} to the female subject in the second image, and {\tt [v3]} to the background in the second image. Where appropriate these tokens could be initialized with CLIP text embeddings semi-representative of the concept ({\em e.g.}, ``person" for both {\tt [v1]} and {\tt [v2]}) or simply initialized with random vector embeddings. The latent masks are then initialized by averaging cross-attention maps (see left of Step 1) between the newly defined tokens and the corresponding images across randomly selected $50$ diffusion timesteps. The resulting attention maps are subsequently binarized and refined using DenseCRF, resulting in latent binary masks (see right of Step 1). 
In Step 2, the tokens are re-optimized for a newly sampled timestep with a combination of Mask Diffusion Loss (3) and Cross-Attention Loss (4) [see text for details]. 
The new tokens are then used to refine cross-attention and the mask. This alternating optimization continues for a fixed number of steps until masks and tokens converge.} 
% and masks are initialized by binarizing the average attention maps over certain timesteps. EM optimization is used to update the tokens given the masks and refine the mask by updating the model weights.}
%\vspace{-0.10in}
\label{fig:modeloverview}
\end{figure*}

\begin{figure*}
%\vspace{-0.10in}
\begin{subfigure}{.325\textwidth}
  \centering
  \includegraphics[scale=.23]{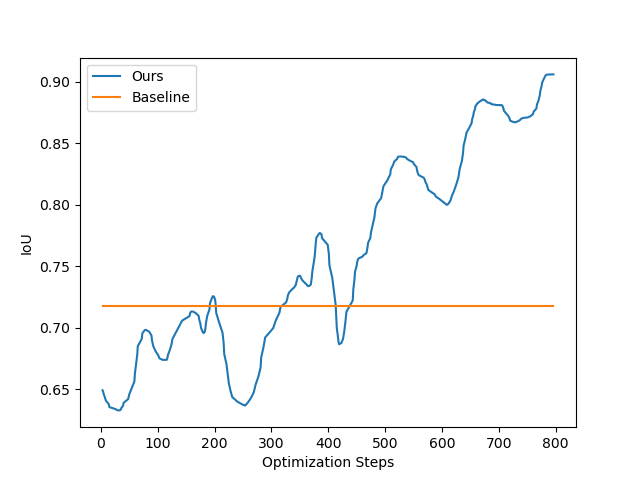}
  \caption{Mask optimization for Token1}
\end{subfigure}
\begin{subfigure}{.325\textwidth}
  \centering
  \includegraphics[scale=0.23]{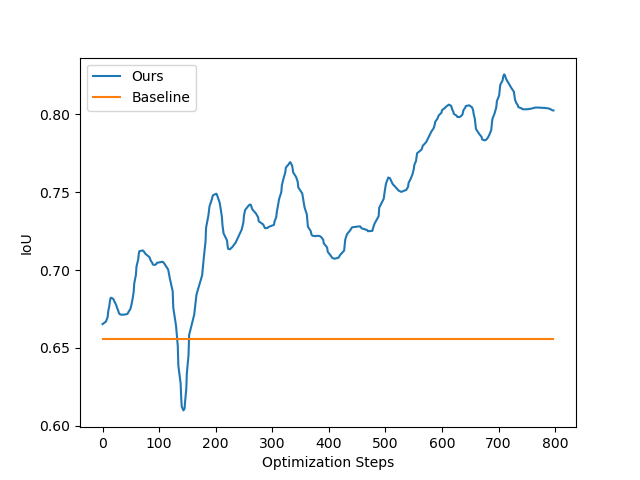}
  \caption{Mask optimization for Token2}
\end{subfigure}
\begin{subfigure}{.325\textwidth}
  \centering
  \includegraphics[scale=0.23]{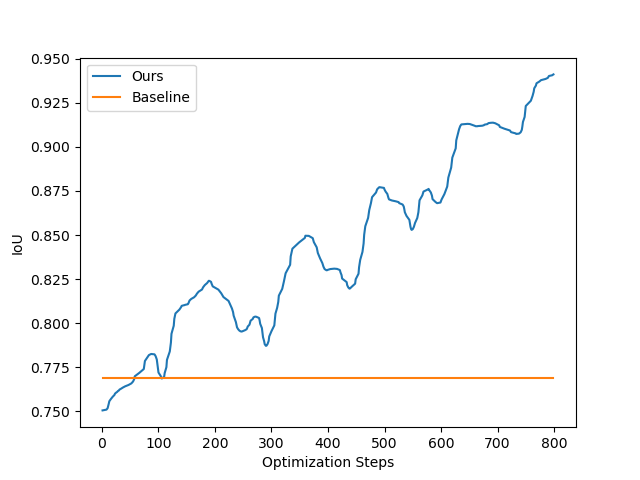}
  \caption{Mask optimization for background Token}
\end{subfigure}
\caption{{\bf  Quantitative comparison - Mask IoU as a function of training steps.} 
We compare the estimated (latent) mask and the manually annotated mask for the concepts illustrated in Figure~\ref{fig1:intro}. 
Performance, as a function of mask optimization, for different tokens over training steps is illustrated: (a) Mask optimization for {\tt [v1]]} (\emph{i.e. subject}), (b) Mask optimization for {\tt [v2]} (\emph{i.e. another subject}), and (c) Mask optimization for {\tt [v3]}.
The yellow line indicates a baseline where the mask is obtained once and not updated jointly with the tokens (hence fixed value). 
% as sampled timesteps in diffusion introduce variance. 
A clear trend of improvement is observed for all masks. Curves are smoothed to highlight trends.
}
%\vspace{-0.15in}
\label{fg:mask_opt}
\end{figure*}
%Mask initialization with diffumask
%Token optimization
%Alternating optimization

To generate images with user-specified concepts (subjects or background) and enable interactions between multiple such concepts and in various contexts, 
 %To solve the insertion of concepts (subjects or background) in target images and enable interactions between multiple subjects, 
it is essential to extract the concept information, disentangle its specific visual representation, and encode it in the diffusion model.
We define a {\em concept} as a visually illustrated instance of an object/subject or a well examplified pattern. Note that such concepts often entangle content and style. 

Diffusion models encode the textual guidance for accurate image generation through cross-attention between the textual tokens and the corresponding region in the image.
Therefore, to associate {\em new} tokens with particular concepts, a masked representation of the target image highlighting only the specific concept is inevitably beneficial. 
We consider an alternating expectation maximization (EM) style optimization to jointly learn: the tokens -- encoding the concept-specific information; and the binary masks -- corresponding to each concept of interest.
This methodology provides for accurate reconstruction of the target concept in arbitrary contexts, even when involving complex interactions between the subjects. 
In the following, we first introduce the backbone and the constituent components of our optimization framework.

%\vspace{-0.15in}
\paragraph{Latent diffusion model.}
Latent Diffusion Models (LDM) \cite{rombach2021highresolution} use perceptual image compression to project the original image to a lower-dimensional space.
%An auto-encoder approach is employed such that the original spatial structure of the input image is preserved in the latent space. 
An encoder $E(\cdot)$ maps the input image $\mathbf{I} \in \mathbb{R}^{H \times W \times 3}$ to a latent representation $\mathbf{Z} \in \mathbb{R}^{h \times w \times c}$, downsampling the image to a lower spatial dimension.
The diffusion model is then applied to the latent $\mathbf{Z}$, where a time-conditioned U-Net $\epsilon_\theta(\mathbf{Z}_t,t)$ models the diffusion process.
The objective of the latent diffusion model is, 
\begin{align}
\mathcal{L}_{LDM} \coloneqq \mathbb{E}_{t,\mathbf{Z},\epsilon}\left[\|\epsilon - \epsilon_\theta(\mathbf{Z}_t,t)\|_2^2\right].
\label{eq:LDM:obj}
\end{align}
%During training, a forward diffusion process is applied to generate $\mathbf{Z}$, which are mapped to the original image space using a decoder $D(\cdot)$.

\paragraph{Mask extraction.}
Following \cite{wu2023diffumask}, we leverage the cross-attention maps to generate the mask of the target concept. 
Given a conditioning prompt $\mathbf{y} \in \mathbb{R}^{N \times d}$ where $N$ is the number of tokens in the prompt and $d$ is the embedding dimension, a cross-attention layer with key $(\mathbf{K})$, query $(\mathbf{Q})$ and value $(\mathbf{V})$ is given by,
\begin{align}
     \mathbf{A} = \text{softmax}\left(\frac{\mathbf{Q}\mathbf{K}^T}{\sqrt{d}}\right) \cdot \mathbf{V}.
    \label{eq:LDM:attention}
\end{align}
The attention map $\mathbf{A}$\footnote{Here, $\mathbf{Q} = \mathbf{W}_{Q}.\hat{f}(\mathbf{Z})$, $\mathbf{K}=\mathbf{W}_K.f(\mathbf{y})$ and $\mathbf{V}=\mathbf{W}_V.f(\mathbf{y})$, and $\mathbf{W}_Q \in \mathbb{R}^{d \times d_q}$, $\mathbf{W}_K \in \mathbb{R}^{d \times d_k}$ and $\mathbf{W}_V \in \mathbb{R}^{d \times d_v}$ are learnable parameters, $\hat{f}(\mathbf{Z})$ an intermediate flattened feature representation of $\mathbf{Z}$ within the diffusion model and $f(\mathbf{y})$ the feature representation of the condition $\mathbf{y}$.} can be reshaped to ${H \times W \times N}$ such that each slice $\mathbf{A}_p \in \mathbb{R}^{H \times W}$ represents the region attended by the $p^{th}$ token.
These attention maps are averaged across all layers of the U-Net to get the response for the regions attended by the target tokens. Note that the latent diffusion objective and hence the attention map are a function of the timestep $t$. Therefore the quality of the mask would in general differ depending on one or a range of timesteps considered. We will discuss this in more detail later.

To obtain the binary mask $\mathbf{B}$, we follow the simple process of using a fixed threshold and assigning a value of 1 for pixel values larger than the threshold and 0 otherwise. 
We further refine the binary masks using the dense conditional random field (DenseCRF) \cite{KrahenbuhlK11}, where 
a DenseCRF, or fully connected CRF, endows pairwise potentials for each pair of pixels in an image where the local relationship, which is based on the 
color and distance of pixels. 
The goal of CRF optimization is to satisfy the soft unary constraints obtained by passing a stack of binary masks for each image through a {\tt softmax}, while attempting to ensure nearby and similar pixels are assigned the same class label.
It yields refined segmentation that respects natural segment boundaries in the image, which we call $\mathbf{M}$. 
% can be harnessed for creating masks from attention maps in \cref{eq:LDM:attention}.
% We apply the denseCRF to get refined masks $\mathbf{M}$ from the binary masks $\mathbf{B}$.
We adopt DenseCRF hyperparameters from CutLER \cite{wang2023cut}.
% without tuning.

\begin{figure*}
  \centering
  \begin{subfigure}{\textwidth}
       \includegraphics[width=\textwidth]{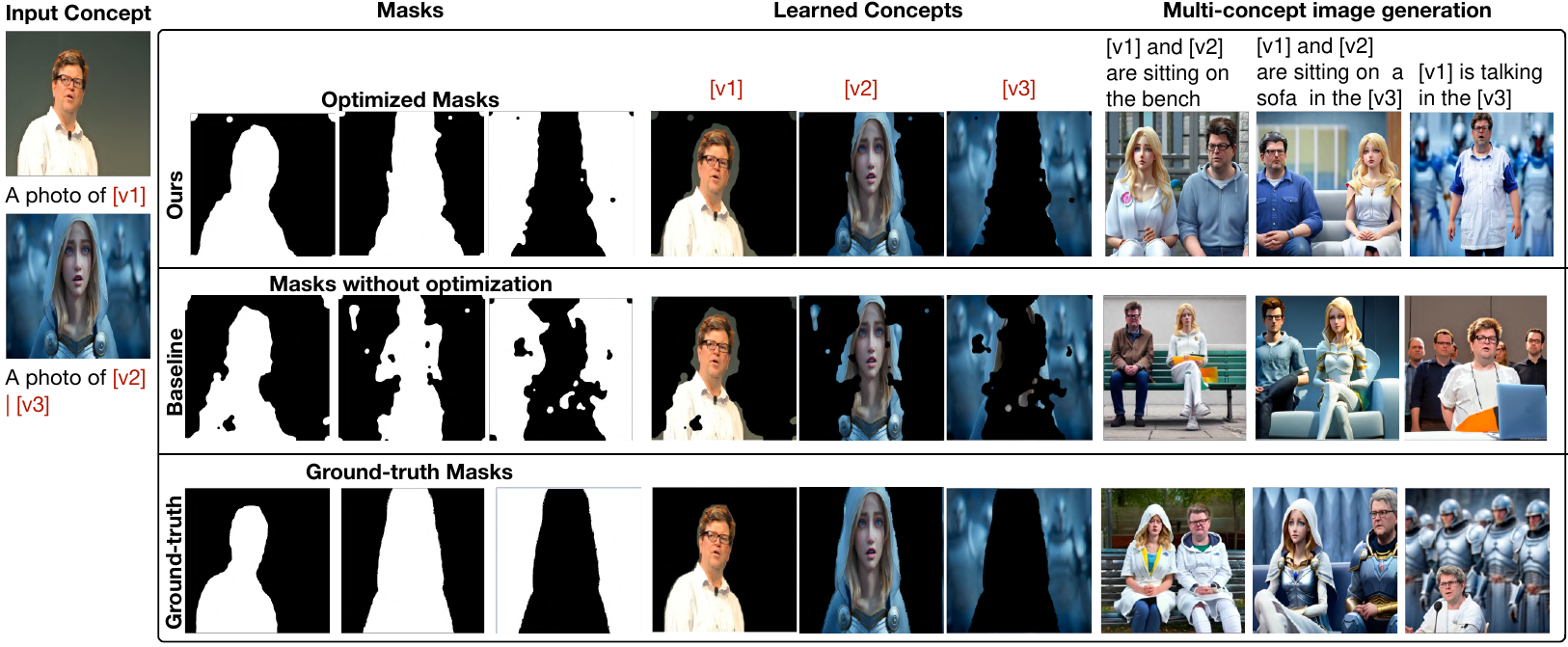}
  \end{subfigure}\\
  %\begin{subfigure}{\textwidth}
      % \includegraphics[width=\textwidth]{sec/figure/qualitative_fig4_2_compressed.pdf}
  %\end{subfigure}
  \caption{{\bf Qualitative results for concept-driven image generation.} Here, we show a comparison between our results with mask optimization, baseline method without mask optimization, and ground truth approach.}
 % \vspace{-0.15in}
  \label{fig:qualatative_results}
\end{figure*}

\begin{figure*}
  \centering
  \includegraphics[width=\textwidth]{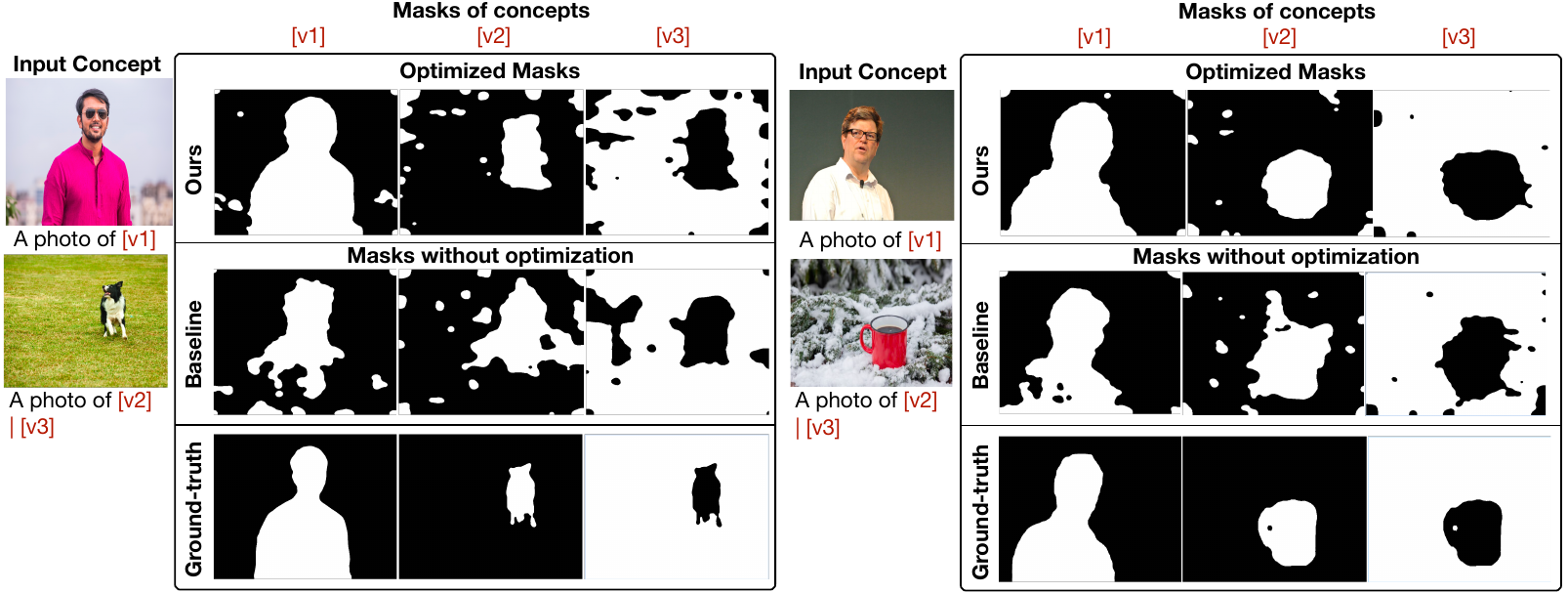}
\caption{
{\bf Comparison of generated masks.} Here we show user-defined masks, masks generated by the baseline method, and those produced by our approach. Note the significant improvement of the mask as a result of joint optimization (first row). }
%\vspace{-0.20in}
\label{fig:mask_gen}
\end{figure*}

\paragraph{Masked diffusion loss.} The input prompt for token optimization is designed such that it contains tokens for target concepts. 
In a prompt $\mathbf{y}$ with $k$ tokens of interest, diffusion loss is applied to the corresponding pixels which in turn are obtained from the binary mask of each token. 
Therefore, considering $\mathcal{M}_k = \cup_{i=1}^{k}\mathbf{M}_i$ to the union of the pixels of $k$ concepts, the masked diffusion loss \cite{avrahami2023break} is given by, 
\begin{align}
\mathcal{L}_{Mask} \coloneqq \mathbb{E}_{t,\mathbf{Z},\epsilon}\left[\|\epsilon \odot \mathcal{M}_k - \epsilon_\theta(\mathbf{Z}_t,t) \odot \mathcal{M}_k\|_2^2\right].
\label{eq:masked-LDM:obj}    
\end{align}

\begin{figure*}[t]
  \centering
  \includegraphics[width=\textwidth]{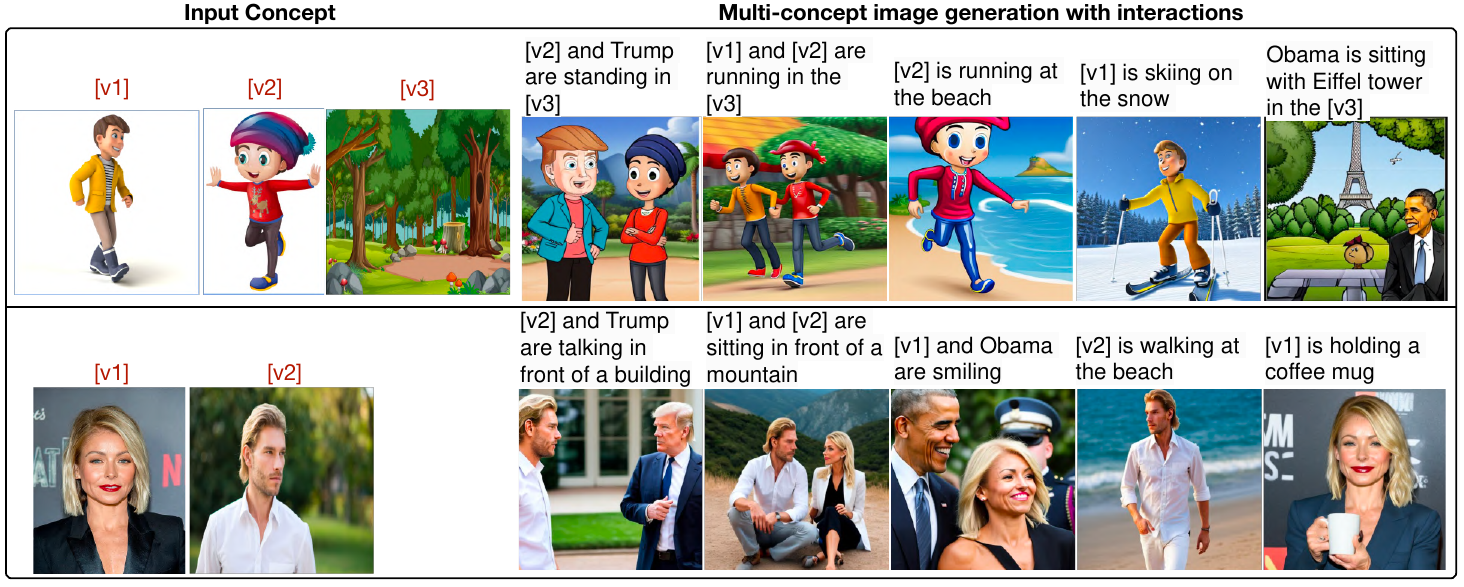}
\caption{
{\bf Interaction between cartoon or real concepts ((including background).} Here, our method is not only able to interact with newly learned concepts but also engage with pre-existing concepts (such as Obama and Trump) in a cartoon-like appearance.}
\label{fig:cartoon}
%\vspace{-0.10in}
\end{figure*} 

\paragraph{Cross attention loss.}
The masked diffusion loss attends to all the desired concept tokens in a prompt. To enforce that a single token encodes information specific to its corresponding instance, we also include the cross-attention loss \cite{avrahami2023break} which encourages the token to attend to only the corresponding target concept,
\begin{align}
\mathcal{L}_{attn} \coloneqq \mathbb{E}_{i,\mathbf{Z},t}\left[\|C_{attn}(\mathbf{Z}_t,\mathbf{y}_i) - \mathbf{M}_i \|_2^2\right].
\label{eq:cattn-LDM:obj}    
\end{align}
Here, $C_{attn}(.)$ is the cross-attention map between the visual representation $\mathbf{Z}_t$ at timestep $t$ and the token $\mathbf{y}_i$ in prompt $\mathbf{y}$ with $k$ concepts, and  $\mathbf{M}_{i}$ is the (latent)  mask for token $\mathbf{y}_i$.

\paragraph{Optimization.}
To ensure that the tokens learned correspond to the target concept, we introduce binary masks to constrain the loss during prompt optimization and update the tokens under the mask.
We adopt a weakly supervised approach and an alternating optimization procedure for learning the concepts.
An EM optimization leads to better convergence under good initialization of the parameters. 

To obtain a reasonable initial estimate of the token embeddings of specific concept-tokens and their corresponding masks, in the first stage, we perform the following updates: 
We first obtain initial concept token embeddings by either initializing them from CLIP 
or by performing inversion without masking to optimize the tokens for a fixed number of iterations. This provides us with good initial token embeddings. 
With these initial token embeddings at hand, we initialize the masks by following the steps outlined in mask extraction and average the masks over certain timesteps to reduce the errors introduced by stochasticity. 

In the second stage, given good initializations of the tokens and their masks, we perform an alternating update of the tokens given the masks and of the masks themselves.
Using \cref{eq:masked-LDM:obj} and \cref{eq:cattn-LDM:obj}, we first optimize the prompt given the masks for each concept at the iteration. 
In the following step, for the optimized token, we compute the \cref{eq:masked-LDM:obj} 
% and \cref{eq:cattn-LDM:obj} to update the model weights to yield an updated mask.
to update the mask (from cross-attention). 
% The masks are, thus, updated and refined during model weight updates yielding better masks that provide better concept coverage. 
Finally, we optimize the tokens and parameters of the model jointly keeping the mask fixed, similar to \cite{avrahami2023break}, this avoids drift. 

The overall optimization is outlined in \cref{fig:modeloverview}. In this example, tokens \texttt{[v1]}, \texttt{[v2]}, and \texttt{[v3]} are assigned to the two subjects and a background of the second image respectively. With a good initialization of the token embeddings, we obtain the initial masks of the three concepts by averaging the attention maps over certain timesteps. Following this, we apply $\mathcal{L}_{Mask}$ and $\mathcal{L}_{attn}$ to update the tokens constrained by the given masks. 
With the newly updated tokens, we update the model weights to get an updated, better-quality mask. We repeat this process over a fixed number of steps (usually 400).
At each iteration, we sample the timesteps to use for the update, similar to \cite{avrahami2023break}.

In this way, our approach can disentangle and learn concepts from a single image with masks generated during optimization.
As illustrated in \cref{fg:mask_opt}, the IoU of our optimized masks with respect to the ground-truth masks consistently increases for different concepts compared to the masks obtained from baseline (without optimization). 
Our approach thus provides flexibility to extract complex concepts without the requirement of a user-defined ground-truth mask.

\section{Experimental Results}

In this section, we will show the experimental results through both quantitative measurements and qualitative assessments. Additionally, we conduct comparisons with baselines for a comprehensive evaluation.

\begin{figure*}
  \centering
  \includegraphics[width=
  \textwidth]{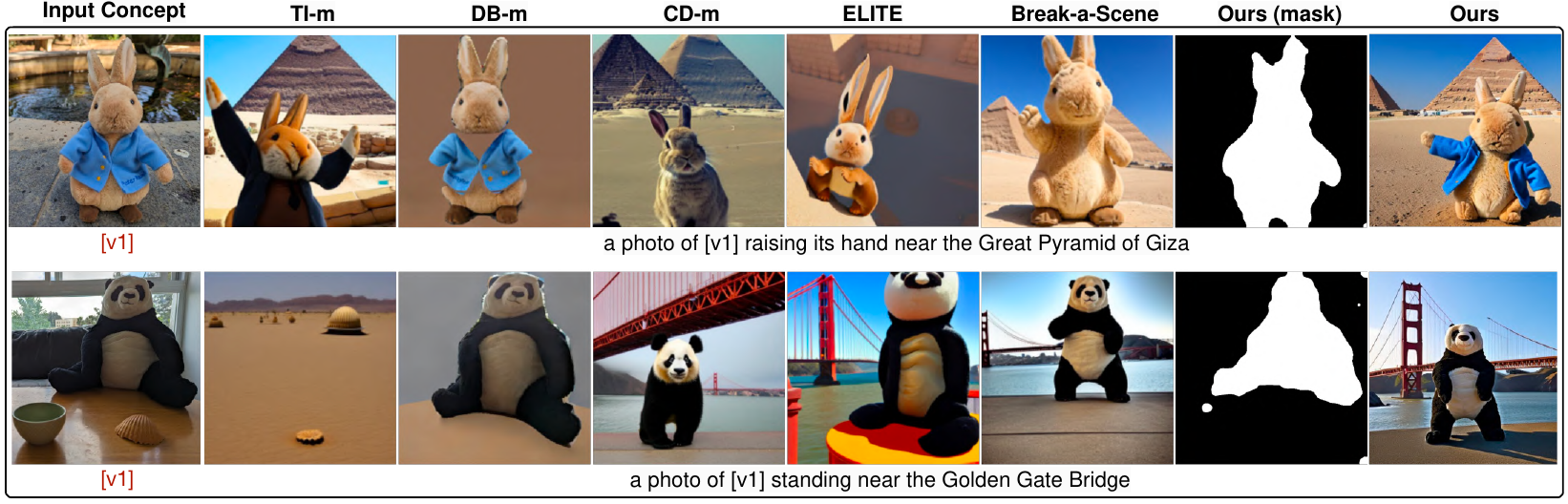}
\caption{
{\bf Comparison of our method with standard baselines.} Here we consider TI-m~\cite{gal2022image}, DB-m~\cite{ruiz2023dreambooth}, CD-m~\cite{kumari2023multi}, ELITE~\cite{wei2023elite}, Break-a-scene~\cite{avrahami2023break} as standard baseline methods and all images are taken directly from ~\cite{avrahami2023break}.}
\label{fig1:rabbit}
\end{figure*}

\begin{figure*}
%\vspace{-0.15in}
  \centering
  \includegraphics[width=\textwidth]{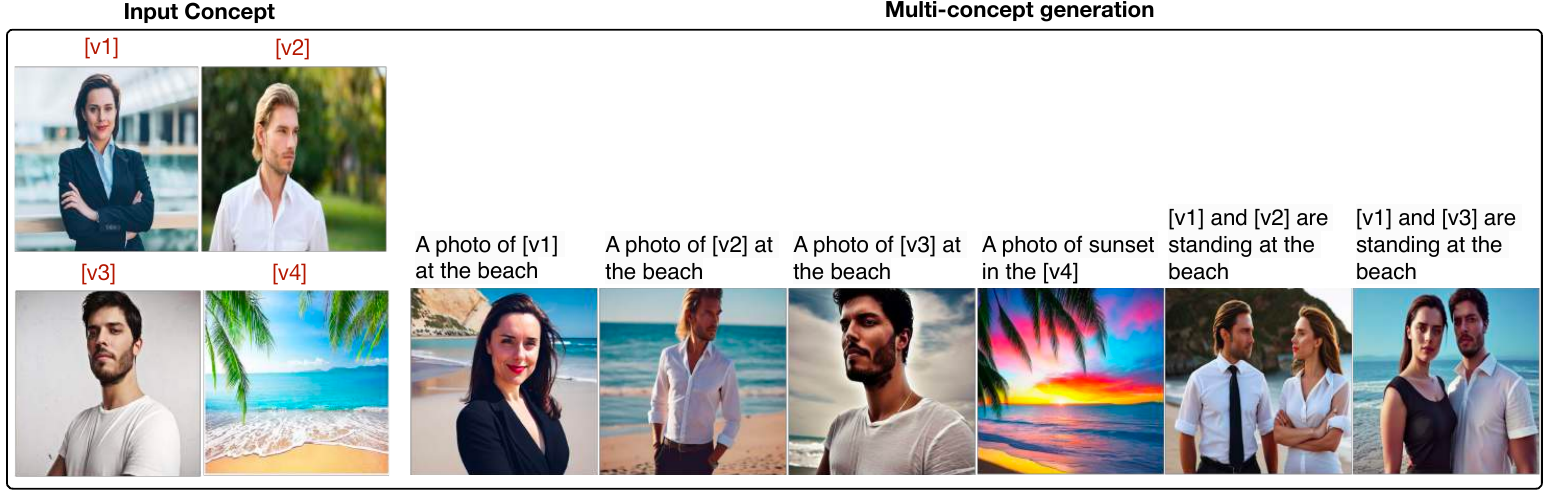}
\caption{
{\bf Qualitative results show that our model can handle four concepts simultaneously.}}
\label{fig1:4_concepts}
%\vspace{-0.15in}
\end{figure*}

\begin{figure*}
%\vspace{-0.17in}
  \centering
  \includegraphics[width=\textwidth]{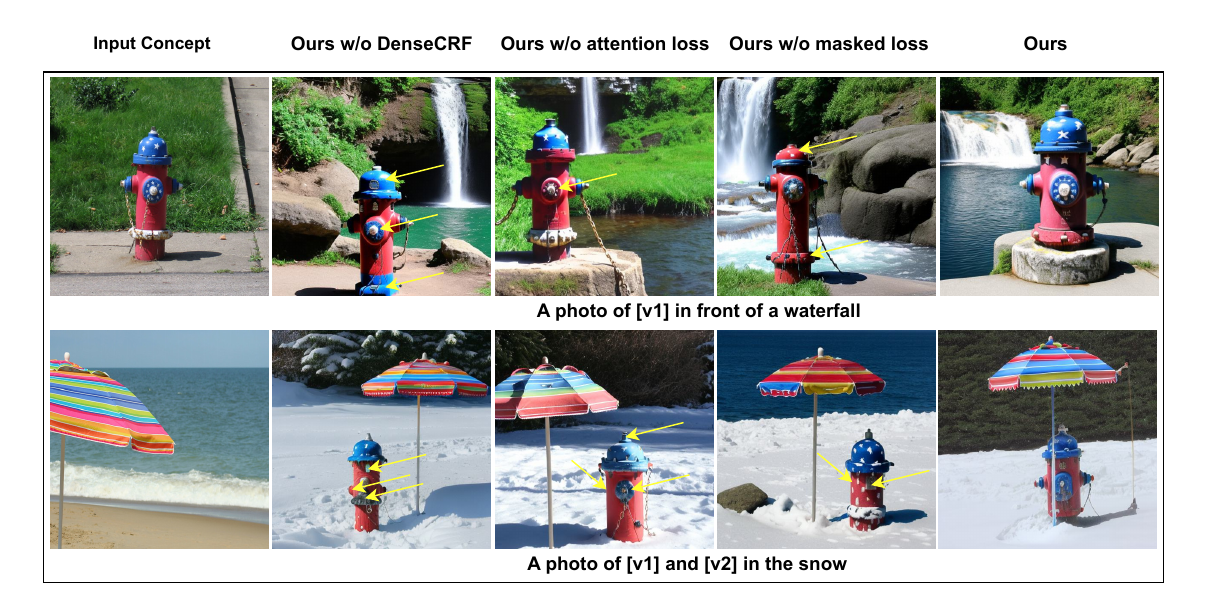}
  %\vspace{-0.20in}
\caption{\textbf{Qualitative ablation by removing different components.}}
\label{fig1:each_component_ablation}
%\vspace{-0.25in}
\end{figure*}

\vspace{0.05in}
\noindent
\textbf{Baseline Methods.} We consider two baselines for comparing our proposed subject-driven image generation method, both of which are subsets of the Break-A-Scene ~\cite{avrahami2023break}. 
These subsets involve: (1) utilizing user-defined masks (\emph{i.e. ground truth masks}) and (2) operating without user-defined masks, which is a weakly supervised baseline (\emph{baseline method}) to compare with our method. 
In this specific baseline approach, we create initial masks by aggregating attention maps derived from a pre-trained stable diffusion~\cite{rombach2021highresolution} model using 50 timesteps, rather than relying on ground truth masks. 
However, there is no optimization of these initial masks in the baseline method. 
We note that (1), which uses human-annotated perfect masks to delineate concepts, is a natural upper bound for our method. That said, it is possible in certain cases for us to outperform it as full-mask may not be optimal for learning the concept in question; the learned (latent) mask allows added flexibility in this respect.

\begin{table}
%\vspace{-0.10in}
\begin{center}
\scriptsize
\caption{\label{tab:table2} \textbf{Quantitative evaluation in terms of image similarity and diversity on COCO dataset.} }
%\vspace{-0.15in}
\begin{tabular}{lccc}%{\textwidth}{@{}Xccc@{}}
\toprule
 Method & Cosine similarity($\uparrow$) & Clip similarity($\uparrow$) &  LPIPS diversity ($\uparrow$) \\ \midrule
Baseline & 0.8245 & 0.8987 & 0.3972  \\ %\midrule
Ours (w/ aug) & 0.81245 & 0.8960 & \bf 0.4134  \\ %\midrule
Ours (w/o aug) & \bf 0.8342 & \bf 0.9062 &  0.40 \\ \bottomrule
\end{tabular}
\end{center}
%\vspace{-0.20in}
%71.8\% users prefer images generated by our approach.} 
\end{table}

\begin{table}
\begin{center}
\scriptsize
%\vspace{-0.10in}
\caption{\label{tab:table1} \textbf{Quantitative evaluation based on user study.} }
%\vspace{-0.15in}
\begin{tabular}{lccc}%{\textwidth}{@{}Xccc@{}}
\toprule
 Data & Baseline & Neutral &  Ours \\ \midrule
Random & 10.92\% & 17.29\% &  \bf 71.80\% \\ %\midrule
COCO & 21.87\% & 18.69\% &  \bf 59.44\% \\ \bottomrule
\end{tabular}
\end{center}
%71.8\% users prefer images generated by our approach.} 
%\vspace{-0.35in}
\end{table}
%\vspace{0.05in}

\vspace{0.05in}
\noindent
\textbf{Quantitative Comparison.} We consider three evaluation metrics to evaluate the faithfulness of the concept-driven generated images approach to the input concepts, considering two measurement criteria as follows:

\begin{itemize}
\item Similarity of the generated images: We use CLIP~\cite{radford2021learning} and Cosine Similarity~\cite{lahitani2016cosine} scores to evaluate the resemblance of the generated images to the target images in terms of semantic similarity. Given a generated image, we use a pre-trained object detector (\emph{e.g. YOLOv8}~\cite{inui2023detection}) to detect the object of interest (corresponding to the input concept) in the generated image and compare their similarity to the target object from MSCOCO images. Table~\ref{tab:table2} displays the results for the baseline method and our method using CLIP and cosine similarity.

\item Diversity between the generated images: We also consider how diverse the generated images for visual concept-driven generation are for a given prompt by comparing them using LPIPS~\cite{zhang2018unreasonable}  across different samples in the Table~\ref{tab:table2}.
\end{itemize}

In Table~\ref{tab:table2}, we present results on the COCO dataset~\cite{lin2014microsoft}.
We randomly selected 120 concepts from 11 COCO object classes, and generated them in various settings using 20 prompts each, creating 8 images for each prompt.
We consider our approach with and without data augmentation. 
Data augmentation makes the generated images more diverse, but it can lead to lower similarity scores when compared to the target images. 
To account for this, similar to the baseline, we also evaluate our method without data augmentation and calculate semantic similarity using CLIP and cosine similarity scores. 
Our method without augmentation still performs better in diversity, due to EM-like optimization, allowing it to generate a wider variety of images compared to the baseline. 
This shows the importance of mask optimization in the process of generating masks.

%There are no existing evaluation metrics to asses our proposed concept-driven image generation method. 
%As a result, we found human evaluation to be the best possible way to validate the approach. 
Moreover, we also found human evaluation to be the best possible way to validate the approach. 
We randomly select 10 prompts and generate 6 images for each of them. 
We carried out a forced-choice experiment involving 11 participants to compare the generated images between our method and the baseline approach.

These images were assessed for faithfulness to the target instance and the prompt condition (representing instances, contexts, and interactions) used to generate the images.
As shown in \Cref{tab:table1}, our method was preferred 71.8\% of the time and was considered on par 17.29\%. 
In addition, we conducted another, much larger scale one, using COCO in the same setting as in \cref{tab:table2}. 
In a user study with 17 participants, our approach was chosen 59.44\% of the time, compared to the baseline. %, which was preferred only 21.87\% of the time.
This clearly illustrates the preference for results obtained using the joint token-mask optimization we propose. 

%we consider five use-cases where we consider learning three tokens (\emph{i.e.} two subjects and one background) and for each use case, we generate a certain number of images (in the first column). In columns two, three, and four, we present the count of accurately generated images and express the accuracy as a \%. Out of the total 416 generated images, the ground truth method successfully generates 234 images. In contrast, the baseline method produces 157 images accurately, while our proposed approach achieves accuracy with 194 images. Therefore, the table illustrates that our mask optimization method outperforms the weakly supervised baseline approach and performs comparably to the ground truth method. In figure~\ref{fg:mask_opt}, We include a quantitative comparison of mask Intersection over Union (IoU) concerning various tokens, plotted against training steps using our proposed method and the baseline approach.

%\begin{figure}[t]
%  \centering
%  \includegraphics[width=\textwidth]{sec/figure/real_real_another.pdf}
%\caption{
%{\bf Interaction between real-world concepts.} Our model can construct interactions with real custom subjects.}
%\vspace{-0.15in}
%\label{fig:real-real}
%\end{figure}

\vspace{0.05in}
\noindent
\textbf{Qualitative Comparisons.} 
Figure~\ref{fig:qualatative_results} shows the results of concept-driven image generation using our method, ground-truth (GT)~\cite{avrahami2023break}, and the baseline method.
The bottom rows exhibit the ground-truth masks, the respective learned concepts from two different use cases, and the images generated for interactions with multiple concepts.
The middle rows and the top rows showcase the weakly supervised generated masks using the baseline and our method, respectively.
In comparison to the baseline, our approach yields images that faithfully represent each concept in multi-concept image generation.

%The first two rows exhibit input concepts and their respective ground truth masks from two different use cases. 
%The subsequent rows~3 and 4 showcase weakly supervised generated masks using the baseline and our method, respectively. Finally, the generated images using GT, the baseline, and our method are presented in the 5th, 6th, and 7th rows.
Figure~\ref{fig:mask_gen} shows qualitative comparisons between the user-defined ground truth masks and the masks generated by both the baseline method and our approach.
We can see that our optimized masks outperform the baseline masks without any optimization. 

\vspace{0.05in}
\noindent
\textbf{Additional Results.} 
%Furthermore, 
we demonstrate (in Figure~\ref{fig:cartoon}) that our approach can interact within a cartoon world context as well. 
In this scenario, the pre-existing characters (\emph{e.g.} Obama and Trump) are also depicted in a cartoon-like appearance. Besides, apart from lexical and cartoon concepts, we conduct experiments utilizing real-world concepts and show that our approach can generate multi-concept interactions between pre-existing and new concepts.
We compare the qualitative results of our method in Figure~\ref{fig1:rabbit}. The comparative images are taken directly from Break-a-Scene~\cite{avrahami2023break}. We can see that our method can generate a better-optimized mask without any spatial supervision. Additionally, the generated images are more aligned with specific subjects and adhere more closely to the provided text prompts.

As an additional experiment, we conduct experiments to see if our model can learn more than three concepts (\emph{e.g.}, four concepts), and the qualitative results are shown in Figure~\ref{fig1:4_concepts}. We can see that our method excels in learning individual concepts but encounters challenges in effectively capturing interactions between these concepts. A possible extension of our proposed approach could address this limitation, and improve the approach to capture interactions between different concepts.
Additional results are provided in the \emph{Supplemental}.

\vspace{0.05in}
\noindent \textbf{Ablation Studies.} To assess the contribution of each component, we conduct qualitative ablation by removing DenseCRF refinement, cross-attention loss, and masked diffusion loss in Figure~\ref{fig1:each_component_ablation}. We can see that when we remove one of the components, the model tends to entangle concepts. Similar behavior has been observed in~\cite{avrahami2023break} for masked diffusion and cross-attention loss. 
Moreover, adding DenseCRF refinement improves the generated masks by 5.9\% in terms of IOU.

%\vspace{-0.1in}
\section{Conclusion}
We present an approach for generating multiple subjects with interconnections to facilitate personalized generation, even incorporating entities beyond out-of-distribution. 
Our method involves an alternating optimization method aimed at separating different concepts within a single image by producing masks tailored to distinct target concepts.
We anticipate its role as a foundational element in the evolution of generative AI, further expanding the horizons of creative expression. %  within the field.

%%%%%%%%% REFERENCES
{\small
\bibliographystyle{plain}
\bibliography{main}
}

\end{document}

\title{Visual Concept-driven Image Generation with Text-to-Image Diffusion Model\linebreak -- Supplemental --}

\author{Tanzila Rahman$^{1,3}$ \qquad  Shweta Mahajan$^{1,3}$ \qquad Hsin-Ying Lee$^{2}$ \qquad Jian Ren$^{2}$ \\ 
\qquad Sergey Tulyakov$^{2}$  \qquad Leonid Sigal$^{1,3,4}$\\
\textit{$^1$University of British Columbia} \qquad \textit{$^2$Snap Inc.} \\
\textit{$^3$Vector Institute for AI} \qquad
\textit{$^4$Canada CIFAR AI Chair} \\
%\texttt{trahman8@cs.ubc.ca, my.yang@mail.utoronto.ca,  lsigal@cs.ubc.ca}
}

\maketitle

\setcounter{figure}{9}

\noindent
In this supplement, we elaborate on our novel approach to generate images driven by visual concepts. We also conduct additional experiments to compare our method with standard approaches, present more examples using our technique, and explore the limitations and potential future directions of our proposed method.

\begin{algorithm}
\tiny
\SetNoFillComment
Input: Diffusion model parameters: $\theta$, Input image: $\mathbf{Z}=E(\mathbf{I})$, Initial prompt: $\mathbf{y}$, Prompt embedding: $\mathbf{\hat{e}}$,
Target concept indices : $K$; Learning rates: $\lambda_p$, $\lambda_{\theta}$,
Timesteps for attention and masks: $T$,
Optimization steps: $N$\\

\tcc{Sample C timesteps}
$T \gets \texttt{random}(0,T,C)$\\
\texttt{\\} \\
\tcc{Prompt Initialization}
\For{$ p \in K$}{
    \For{$ t \in T$}
    {
    $g = \nabla_{\mathbf{\hat{e}}_{p}}[\mathcal{L}_{LDM}(\epsilon_\theta(\mathbf{Z}_t,t,\mathbf{\hat{e}}))]$\\
    $\mathbf{\hat{e}}_p = \mathbf{\hat{e}}_p - \lambda g$\\
    }
    }
\texttt{\\} \\
\tcc{Initial Attention Map}
$\mathbf{A}_p \gets \texttt{From \cref{eq:LDM:attention}}$\\
\tcc{Mask Initialization}
\For{$ p \in K$}{
    $\mathbf{B}_p \gets \nicefrac{1}{|T|}\sum_t{\texttt{binarize}((\mathbf{A}_p)_t)}$ \\
    $\mathbf{M}_p \gets \texttt{DenseCRF}(\mathbf{B}_p)$ 
    }
\texttt{\\} \\
\tcc{EM optimization of tokens and masks}
\For {$n \gets 1 $ \KwTo $N $}{
\tcc{Sample C timesteps}
 $T \gets \texttt{random}(0,T,C)$\\ 
\texttt{\\} \\
\tcc{M-step: Optimization} 
\eIf{$n \leq \frac{N}{2}$}{
\tcc{Prompt optimization}
\For{$ p \in K$}{
    \For{$ t \in T$}
    {
    $g = \nabla_{\mathbf{\hat{e}}_{p}}[\mathcal{L}_{mask}(\epsilon_\theta(\mathbf{Z}_t,t,\mathbf{\hat{e}}),\mathbf{M}_p)]$\\
    $\mathbf{\hat{e}}_p = \mathbf{\hat{e}}_p - \lambda_p g$\\
    }
    } 
}{
\tcc{Diffusion parameter optimization}
 \For{$ p \in K$}{
    \For{$ t \in T$}{
    $g = \nabla_{\theta}[\mathcal{L}_{attn}(C_{attn}(\mathbf{Z}_t,t,\mathbf{\hat{e}_p}),\mathbf{M}_p)]$\\
    $\theta = \theta-\lambda_{\theta} g$\\
    }   
    }
} 
\texttt{\\} \\
    \tcc{E-step: Mask computation}
    $\mathbf{A}_p \gets \texttt{Update using \cref{eq:LDM:attention}}$ \\
    \For{$ p \in K$}{
    $\mathbf{B}_p \gets \nicefrac{1}{|T|}\sum_t{\texttt{binarize}((\mathbf{A}_p)_t)}$ \\
    $\mathbf{M}_p \gets \texttt{DenseCRF}(\mathbf{B}_p)$ 
    }
}
return $\mathbf{\hat{e}}$, $\theta$   
 \caption{EM-like Optimization of Prompt Tokens and Masks}
 \label{algo:neg_samp:supp}
\end{algorithm}

\section{Algorithm}
We show an overview of our proposed visual concept-driven image-generation approach in Algorithm~\ref{algo:neg_samp:supp}.  The algorithm presents an optimization process akin to Expectation Maximization (EM) to untangle different elements within images. It achieves this by creating masks that isolate particular desired concepts within the images, while, at the same time, learning tokens to represent them. This process of optimization not only improves the learning of more effective concept tokens but also yields enhanced latent masks for specific visual concepts.

\begin{figure*}[t!]
  \centering
 % \begin{subfigure}{\textwidth}
      % \includegraphics[width=\textwidth]{sec/figure/qualitative_fig4_1_compressed.pdf}
 % \end{subfigure}\\
  \begin{subfigure}{\textwidth}
      \includegraphics[width=\textwidth]{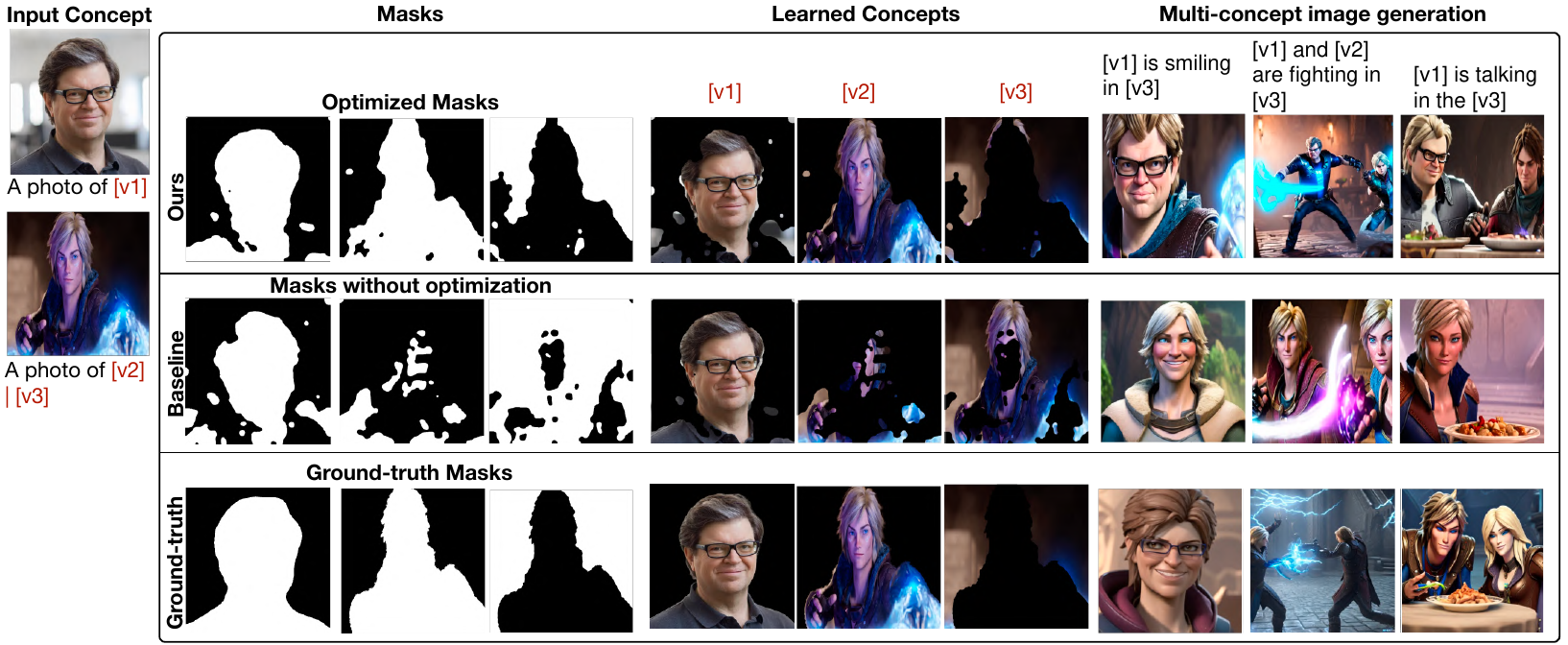}
  \end{subfigure}
  \caption{{\bf Qualitative results for concept-driven image generation.} Here, we show a comparison between our results with mask optimization, baseline method without mask optimization, and ground truth approach.}
 % \vspace{-0.15in}
  \label{fig:qualatative_results_}
\end{figure*}
\begin{figure*}[h!]
  \centering
  \includegraphics[width=\textwidth]{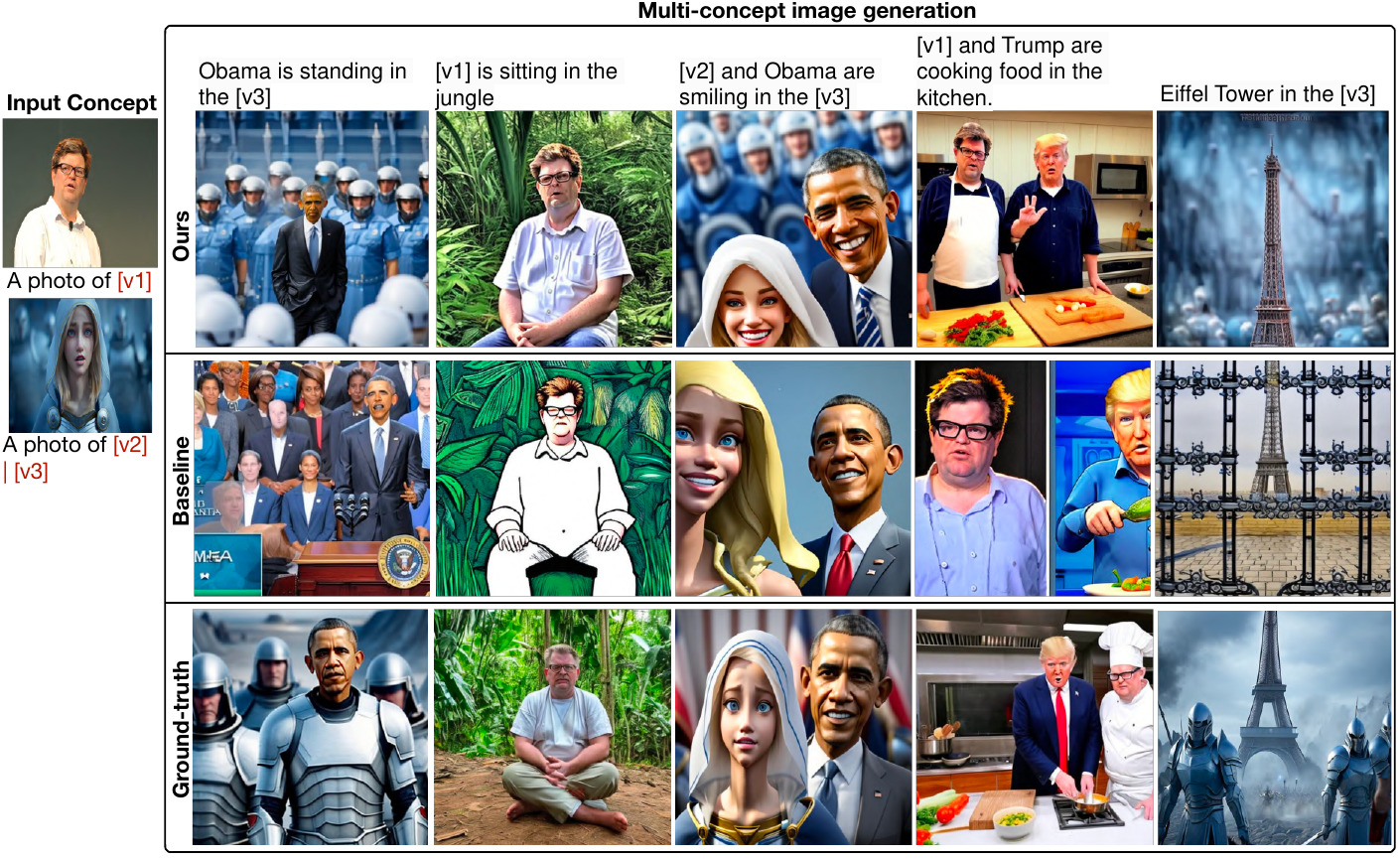}
  \caption{{\bf Qualitative results demonstrating the interaction between different concepts.} Here we present the interaction between newly learned concepts and the pre-existing concepts already familiar to the stable diffusion model.}
 % \vspace{-0.25in}
  \label{fig:in_out_distibution}
\end{figure*}

\section{Additional Qualitative Results}
\begin{figure*}
  \centering
 \includegraphics[width=\textwidth]{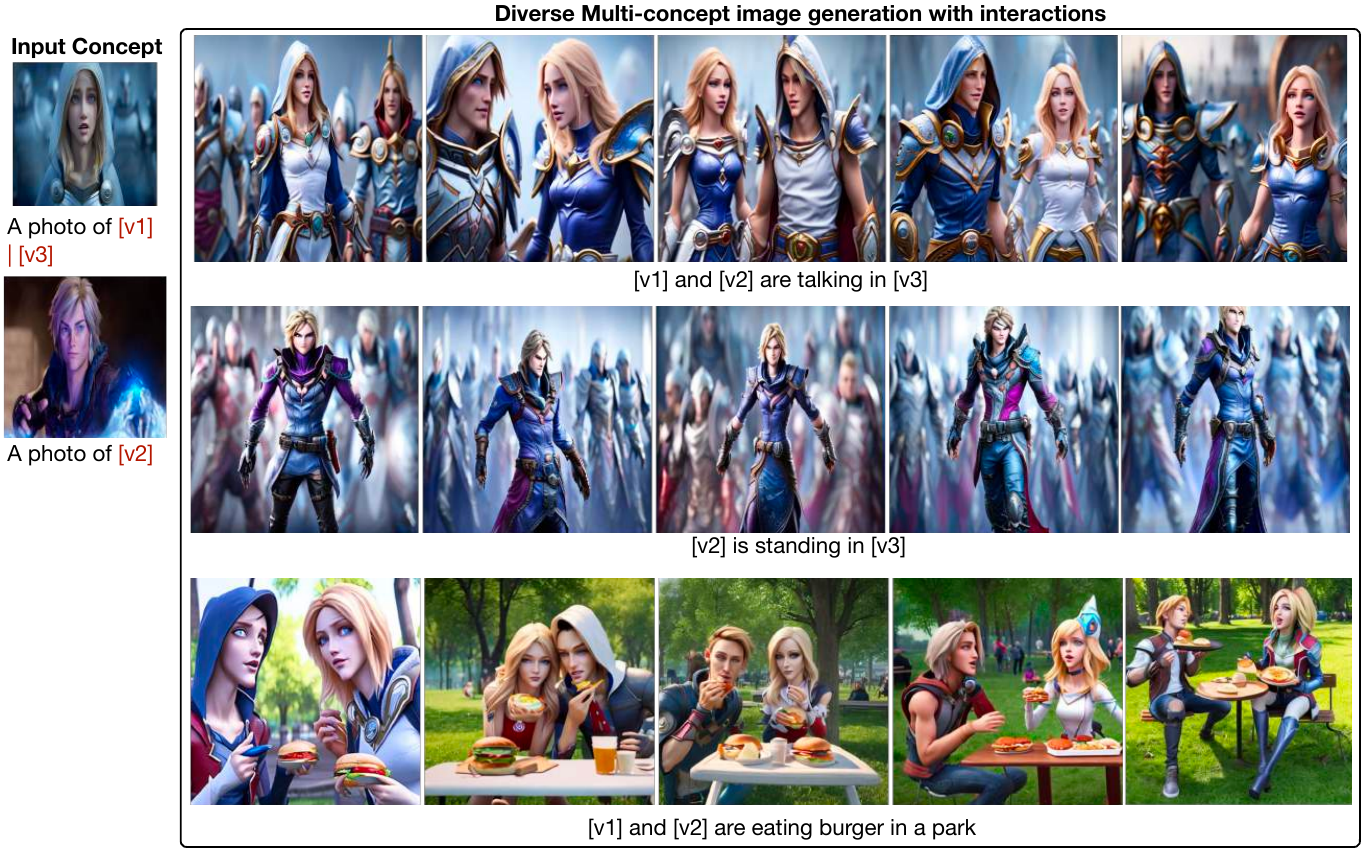}
\vspace{-0.25in}
\caption{
{\bf Diverse images for the same prompt.} Given input concepts, our method demonstrates the ability to generate a variety of diverse outputs for a single prompt. }
\label{fig:diversity}
\vspace{-0.15in}
\end{figure*}

Figure \ref{fig:qualatative_results_} shows additional results to compare our results with mask optimization, baseline method without mask optimization, and ground-truth approach.
In Figure~\ref{fig:in_out_distibution}, we present the comparative results to show the interaction between newly introduced concepts and the pre-existing ones within the stable diffusion model. 
Our proposed method is not only able to interact between multiple concepts but also shows its capability to produce diverse outputs for a single prompt, as illustrated in Figure~\ref{fig:diversity}. 

\begin{figure*}
  \centering
  \includegraphics[width=\textwidth]{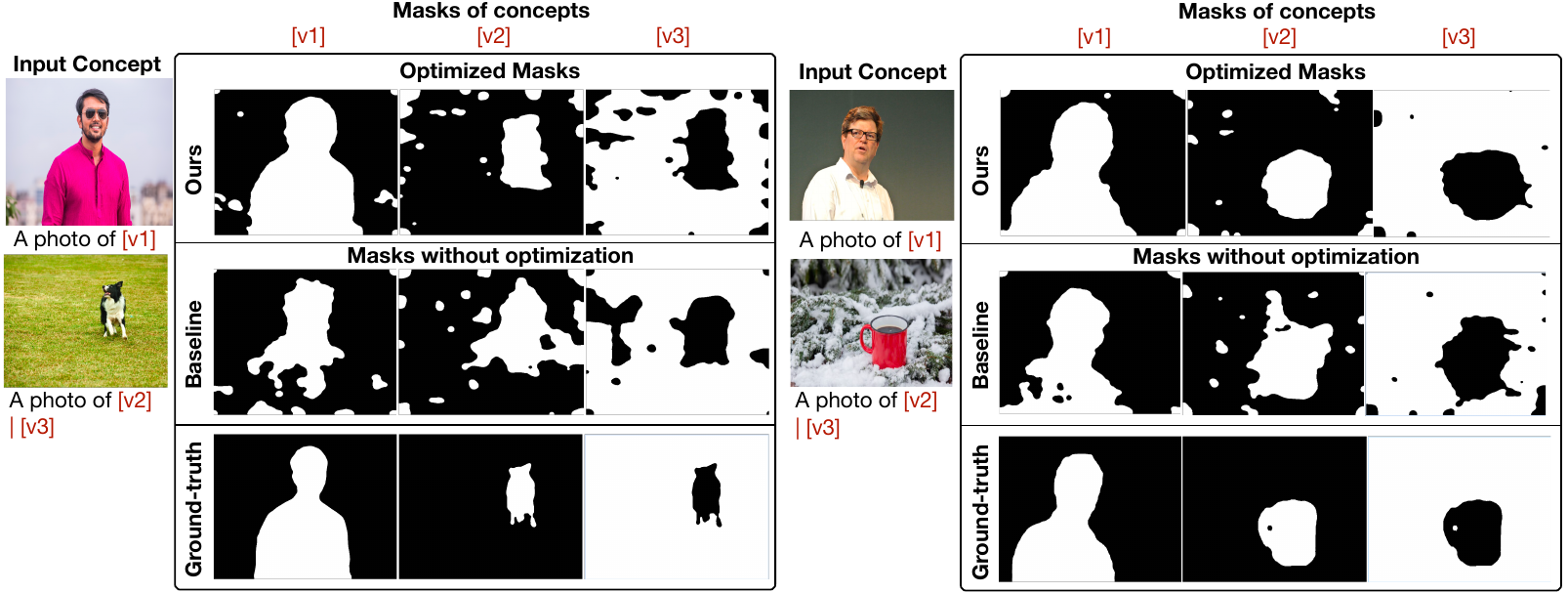}
\caption{
{\bf Comparison of generated masks.} Here we show user-defined masks, masks generated by the baseline method, and those produced by our approach. Note the significant improvement of the mask as a result of joint optimization (first row). }
%\vspace{-0.20in}
\label{fig:mask_gen}
\end{figure*}

Figure~\ref{fig:mask_gen} shows qualitative comparisons between the user-defined ground truth masks and the masks generated by both the baseline method and our approach.
We can see that our optimized masks outperform the baseline masks without any optimization. 

\begin{figure*}
  \centering
  \includegraphics[width=\textwidth]{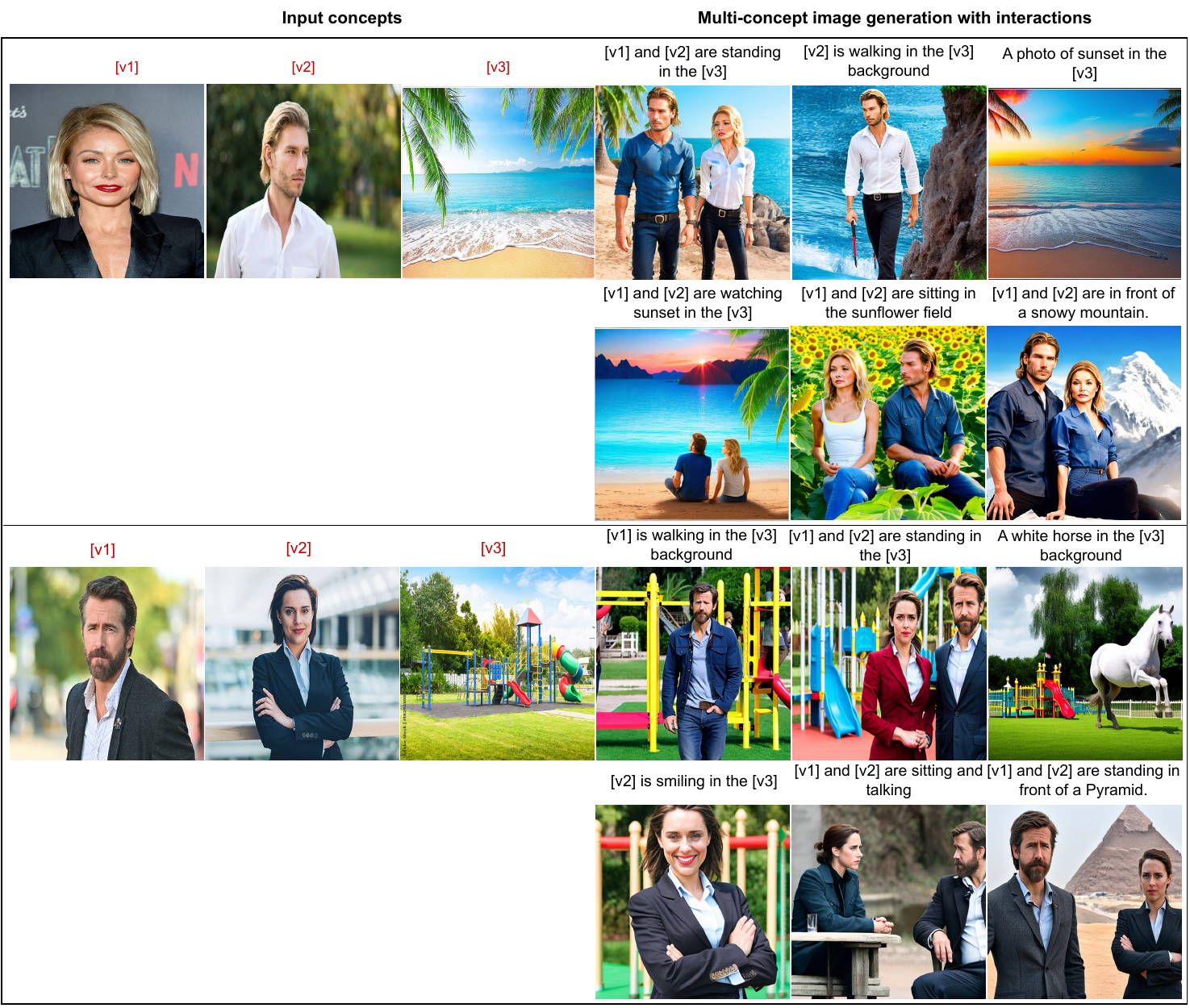}
\caption{
{\bf Interaction between real-world concepts (including background).} Our model demonstrates the capacity to generate accurate interactions among real-world concepts.}
\label{fig1:real_real_sup}
\end{figure*}

\begin{figure*}
  \begin{center}
  %\vspace{-0.25in}
  %\hspace{-.35in}
    \includegraphics[width=0.6\textwidth]{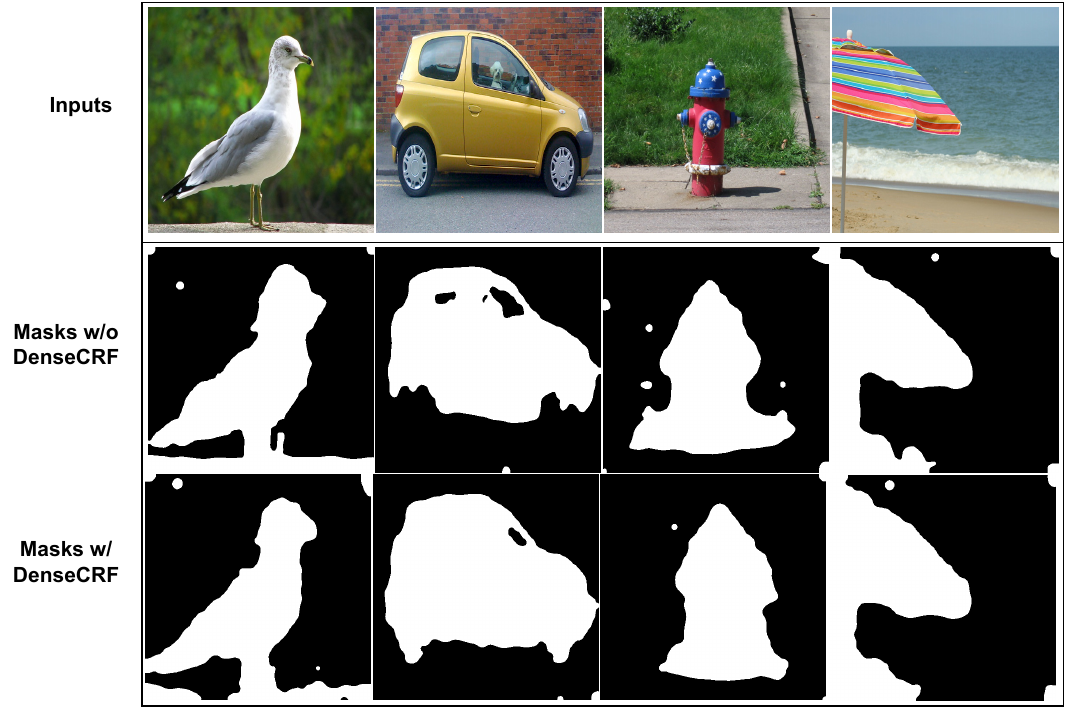}
  \end{center}
  \vspace{-0.18in}
\caption{\textbf{Generated mask w/ and w/o DenseCRF.}}
  \label{fig1:densecrf_ablation}
  \vspace{-0.20in}
\end{figure*}

\begin{figure*}
  \begin{center}
  %\vspace{-0.25in}
  %\hspace{-.35in}
    \includegraphics[width=\textwidth]{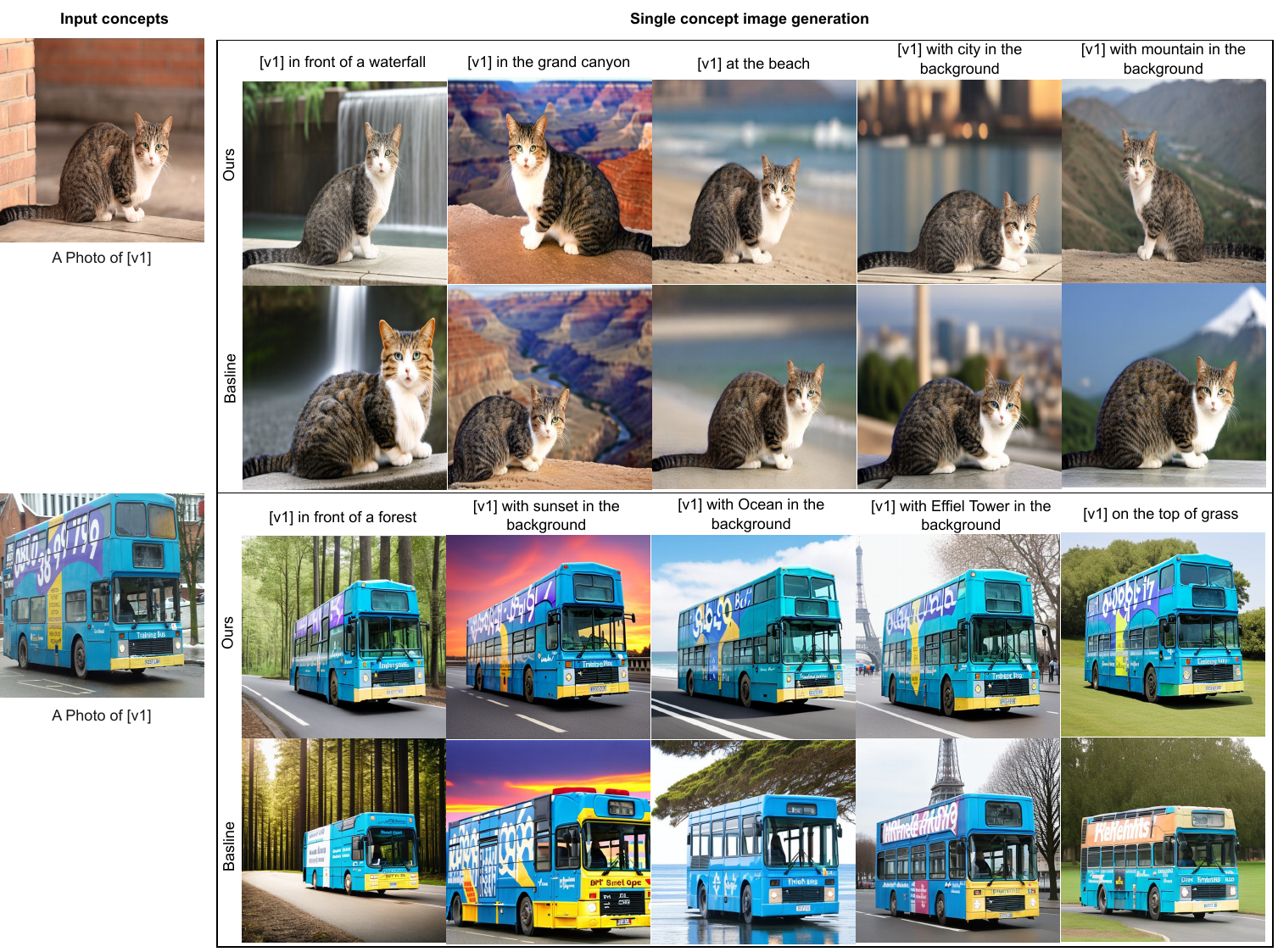}
  \end{center}
  \vspace{-0.18in}
\caption{\textbf{Qualitative results for single concept image generation.} Here, we show a comparison between our method and baseline approach using COCO images.}
  \label{fig1:coco_single_conept}
  \vspace{-0.20in}
\end{figure*}

\begin{figure*}
  \begin{center}
  %\vspace{-0.25in}
  %\hspace{-.35in}
    \includegraphics[width=\textwidth]{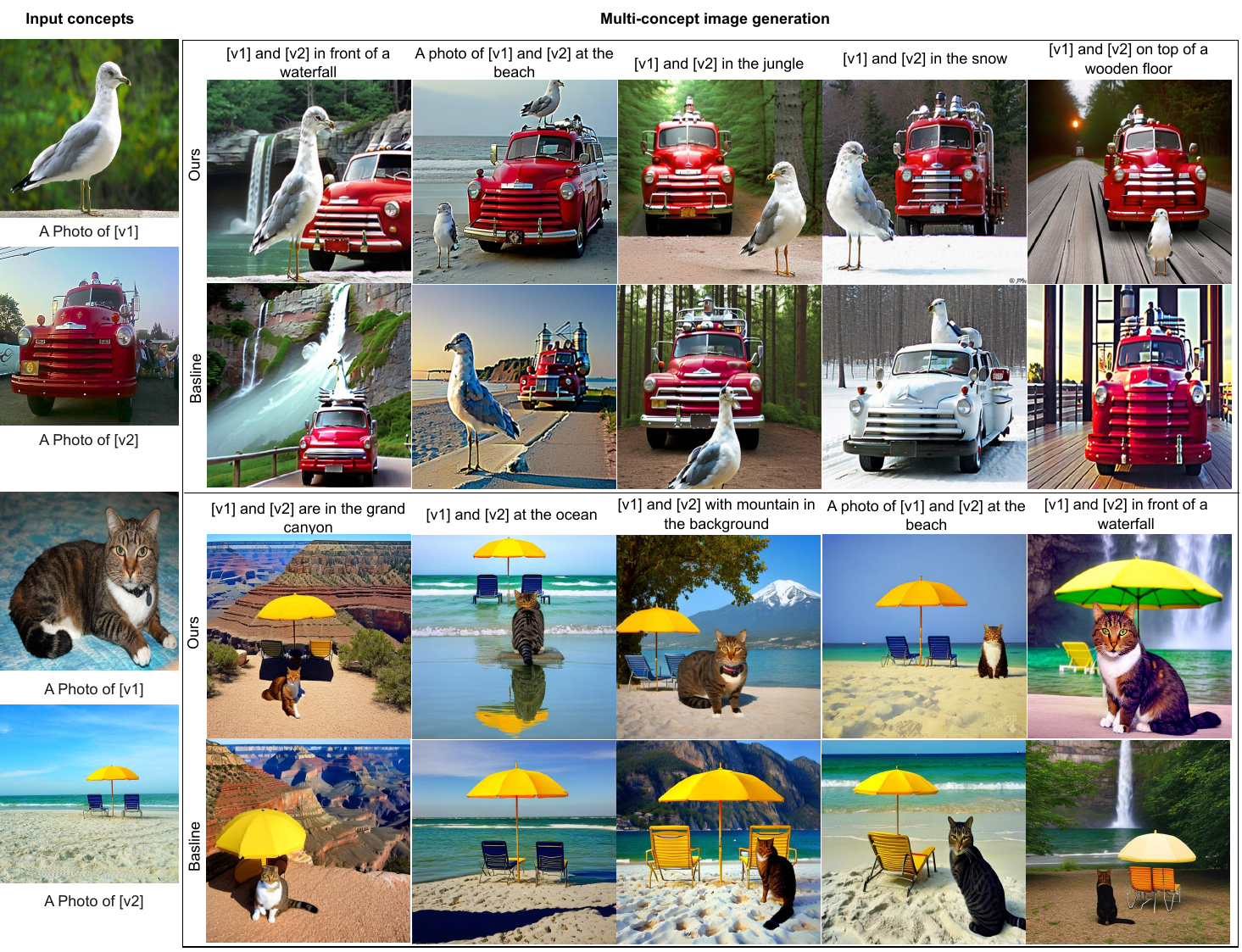}
  \end{center}
  %\vspace{-0.18in}
\caption{\textbf{Qualitative results for multi-concept image generation.} Here, we show a comparison between our method and baseline approach with interactions where input images are taken from COCO dataset.}
  \label{fig1:coco_multi_conept}
  %\vspace{-0.20in}
\end{figure*}

Furthermore, In Figure~\ref{fig1:real_real_sup}, we also present qualitative results that demonstrate our model's ability to engage with real-world concepts. Moreover, we show the importance of adding DenseCRF refinement while generating masks in Figure~\ref{fig1:densecrf_ablation}. Looking at the Figure, it is evident that the DenseCRF refinement significantly enhances the fine details and boundaries in the resulting masks. Refining the initial predictions and taking into account pixel-level interactions and dependencies, leads to a more accurate and visually appealing segmentation mask. In addition, we also present qualitative results where we learn about both single and multi-concepts from the COCO \cite{lin2014microsoft} images in Figures ~\ref{fig1:coco_single_conept} and \ref{fig1:coco_multi_conept} respectively. We also compare the results generated by our method with the baseline method. From the figures, it is clearly visible that our method is able to maintain fine details about the visual concept compared to the baseline method.

%As an additional ablation, we conduct experiments to see if our model can learn more than three concepts (\emph{e.g.}, four concepts), and the qualitative results are shown in Figure~\ref{fig1:4_concepts}. We can see that our method excels in learning individual concepts but encounters challenges in effectively capturing interactions between these concepts. A possible extension of our proposed approach could address this limitation, and improve the approach to capture interactions between different concepts.

\section{Ablation Studies}

\begin{figure*}
  \centering
  \includegraphics[width=\textwidth]{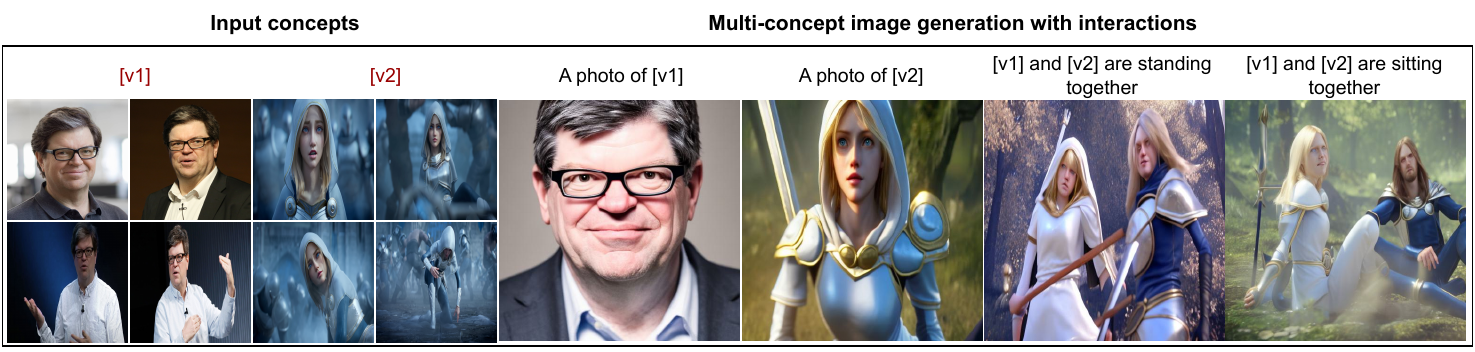}
\caption{
{\bf Utilization of multiple images for a single concept.} The model can grasp individual concepts well but faces difficulty in connecting or interacting between these concepts.}
\label{fig1:multiimages}
\end{figure*}

\noindent \textbf{Runtime cost:} The additional cost during {\em training} is just 0.5\% more than the baseline method and 0.63\% more than the break-a-scene.
The {\em inference} cost remains the same across all models.

%\noindent \textbf{Diversity evaluation:} To measure the diversity of the generated images, we use LPIPS. Our method yields LPIPS of 0.68, while that of baseline method is 0.66. These values indicate that our method produces more diverse images compared to the baseline method.

\noindent \textbf{Why EM-like optimization: } In our proposed approach, we follow a cyclical process reminiscent of the Expectation-Maximization (EM) algorithm.  The analogy to the EM algorithm implies that the alternating steps contribute to refining the overall optimization process. In EM, the E-step involves estimating the expected values of the hidden variables given the observed data, and the M-step involves maximizing the likelihood of the parameters, incorporating the information from the E-step. Similarly, in our method, the alternating steps likely contribute to improving the model's understanding of both the custom tokens and latent masks, with each iteration refining the knowledge gained from the other.

%In our method, we alternate between learning custom tokens and estimating {\em latent} masks that capture corresponding concepts in user-supplied images.  This strategy reflects a cyclic process akin to the iterative nature of EM, where alternating between E-step (Expectation) and M-step (Maximization) contributes to refining the overall optimization process. 

\iffalse
\begin{figure*}[h!]
  \centering
  \includegraphics[scale=0.65]{sec/figure/5_concepts.pdf}
\caption{
{\bf Qualitative results show that our model can handle five concepts but sometimes is unable to interact between concepts.}}
\label{fig1:5_concepts}
\end{figure*}
\fi

\begin{figure*}
  \centering
  \includegraphics[width=\textwidth]{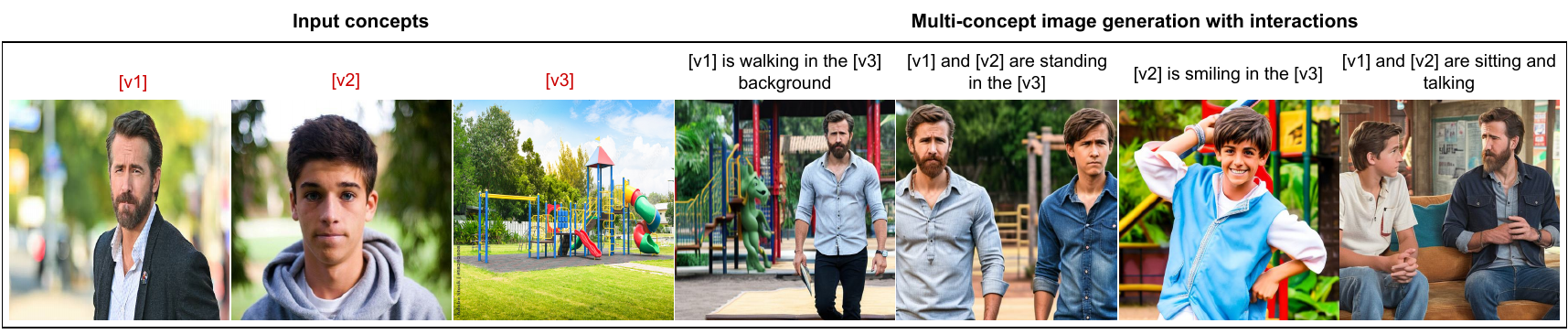}
\caption{
{\bf Stable Diffusion Age Bias.} The model demonstrates a bias towards adult individuals, resulting in a noticeable alteration in the boy's appearance.}
\label{fig1:limitation1}
\end{figure*}
\begin{figure*}
  \centering
  \includegraphics[width=\textwidth]{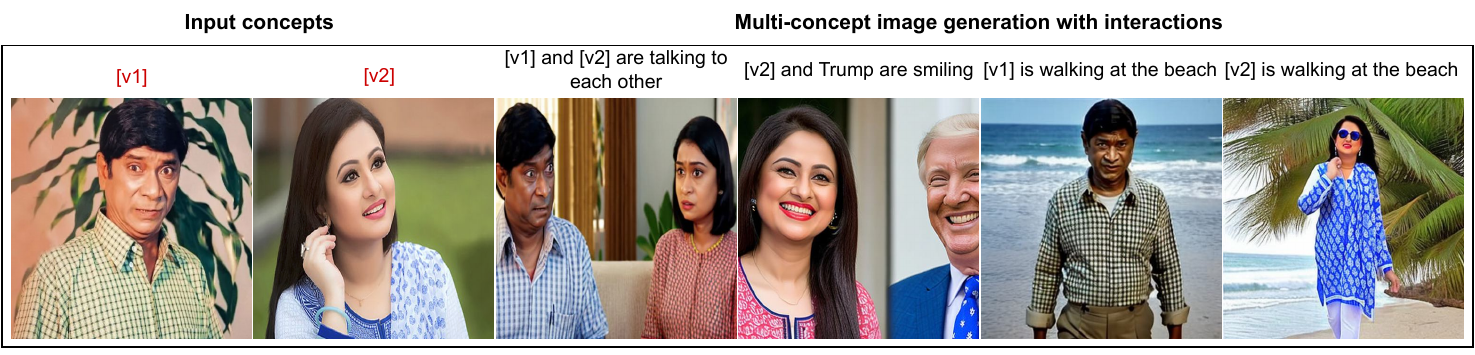}
\caption{
{\bf Stable Diffusion Caucasian Bias.} The model exhibits a stronger bias towards data associated with the Caucasian demographic or region.}
\label{fig:limitation2}
\end{figure*}

\section{Limitations and Future Directions}
Our proposed approach and Break-a-Scene~\cite{avrahami2023break} both employ single images to learn specific concepts. However, considering the utilization of multiple images for a single concept, could this potentially enhance the learning of the concept? The qualitative results for the experiment are shown in Figure~\ref{fig1:multiimages}. We observed that multiple images for a single concept do not enhance the interaction between multiple concepts. Despite the model's improved understanding of each concept, there is no substantial improvement in the overall interaction among concepts. Therefore, it could be a potential future direction to improve the quality of generated images.

In this work, we adopted a pre-trained stable diffusion~\cite{han2023highly} model as the foundation and fine-tuned it to grasp concepts for improved image generation. However, our observations revealed the presence of stable biases, including age and Caucasian bias, as depicted in Figures~\ref{fig1:limitation1} and ~\ref{fig:limitation2} respectively. Working to remove these stable diffusion biases could be another potential research direction to extend the proposed approach.

%%%%%%%%% REFERENCES
{\small
\bibliographystyle{plain}
\bibliography{main}
}